\documentclass{article} 
\PassOptionsToPackage{numbers, compress}{natbib}
\usepackage[main, final]{neurips_2026}
\usepackage{times}

\usepackage{amsmath,amsfonts,bm}

\def\figref#1{figure~\ref{#1}}

\def\secref#1{section~\ref{#1}}

\def\eqref#1{equation~\ref{#1}}

\def\1{\bm{1}}

\def\vzero{{\bm{0}}}

\def\vg{{\bm{g}}}
\def\vh{{\bm{h}}}

\def\vm{{\bm{m}}}

\def\vr{{\bm{r}}}

\def\vv{{\bm{v}}}

\def\vx{{\bm{x}}}
\def\vy{{\bm{y}}}
\def\vz{{\bm{z}}}

\def\mA{{\bm{A}}}

\def\mI{{\bm{I}}}
\def\mJ{{\bm{J}}}
\def\mK{{\bm{K}}}
\def\mL{{\bm{L}}}
\def\mM{{\bm{M}}}

\def\mP{{\bm{P}}}
\def\mQ{{\bm{Q}}}

\def\mT{{\bm{T}}}

\DeclareMathAlphabet{\mathsfit}{\encodingdefault}{\sfdefault}{m}{sl}
\SetMathAlphabet{\mathsfit}{bold}{\encodingdefault}{\sfdefault}{bx}{n}

\def\gD{{\mathcal{D}}}

\def\sC{{\mathbb{C}}}

\def\sR{{\mathbb{R}}}

\newcommand{\ddt}{\frac{\mathrm{d}}{\mathrm{d}t}}
\newcommand{\E}{\mathbb{E}}

\newcommand{\R}{\mathbb{R}}

\newcommand{\diag}{\mathrm{Diag}}

\newcommand{\norm}[1]{\Vert #1 \Vert}
\newcommand{\abs}[1]{\vert #1 \vert}

\usepackage{wrapfig}
\usepackage{hyperref}
\usepackage{url}
\usepackage{algorithm, algpseudocode}
\usepackage{mathtools}
\usepackage{amsthm}
\usepackage{xcolor}

\newlength{\colw}
\usepackage{booktabs}
\usepackage{url}
\usepackage{amsmath}
\usepackage{graphicx}
\usepackage[percent]{overpic}   
\usepackage{placeins}  
\usepackage{tikz}
\usepackage{amssymb}
\usepackage{enumitem}

\newtheorem{theorem}{Theorem}[section]

\newtheorem{definition}[theorem]{Definition} 
 
\newtheorem{example}[theorem]{Example} 
\newtheorem{remark}[theorem]{Remark}

\usepackage[createShortEnv]{proof-at-the-end}

\newtheorem{corollary}[theorem]{Corollary}

\newtheorem{lemma}[theorem]{Lemma}

\newtheorem{proposition}[theorem]{Proposition}

\title{
A Stability Analysis of AdamW: Unstable Equilibria and Non-Convergence
}

\author{%
  Baiyu Su\thanks{Corresponding author: \texttt{baiyusu@utexas.edu}.}\quad
  Lizhang Chen\quad
  Qiang Liu\\
  The University of Texas at Austin
}

\begin{document} 

\maketitle 

\begin{abstract}
AdamW is ubiquitous in deep learning, yet its behavior remains poorly understood. We analyze its dynamics through the lens of dynamical systems and show that AdamW admits an \emph{implicit fixed-point objective}: its fixed points coincide with the stationary points of a constrained and regularized optimization problem. However, not all of these fixed points are stable under AdamW’s dynamics, and stability depends sensitively on curvature, weight decay, and momentum parameters. Even in one-dimensional settings, equilibria may be unstable, and trajectories can exhibit persistent oscillations rather than convergence to them. We further extend the analysis to higher dimensions, deriving sufficient conditions for local stability of the continuous-time dynamics, and propose a convergent modification that preserves AdamW's fixed points. These results clarify what optimization problem AdamW is associated with, when its convergence can be expected, and how its dynamics could inspire more reliable optimizers.
\end{abstract}

\section{Introduction} \label{sec:introduction}

The Adam optimizer with decoupled weight decay (AdamW; \citep{loshchilov2017decoupled}) is the default optimizer across modern deep learning systems, including large language models, vision architectures, and generative models \citep{grattafiori2024llama, liu2024deepseek, ravi2024sam, esser2024scaling}. Yet beneath this ubiquity lies a fundamental gap: while the convergence of Adam (without weight decay; \citep{kingma2014adam}) is already subtle to analyze, \emph{no general convergence proof is available for AdamW}\footnote{See \secref{sec:related_works} for discussion.}. The difficulty arises from the use of decoupled weight decay, which is essential in practice for stability and generalization \citep{kosson2023rotational, d2024we}, but remains disconnected from the objective, misaligned with the gradient, and fundamentally alters the algorithm’s dynamics.  
   
\paragraph{Our contributions.}
First, we show that AdamW can fail to converge even on simple static objectives: its fixed points are exactly the stationary points of an implicit objective, but not all are stable. We give exact stability conditions in one dimension and sufficient ones in higher dimensions, with counterexamples for both $\beta_1>\beta_2$ and $\beta_1<\beta_2$ (sections~\ref{sec:objective}--\ref{sec:multidimensional}). Second, we propose AdamMe, a rescaled weight decay that preserves AdamW's fixed points and admits a Lyapunov function guaranteeing convergence of bounded continuous-time trajectories (\secref{sec:adamme}).

\begin{figure}[t]
    \centering
    \includegraphics[width=\textwidth]{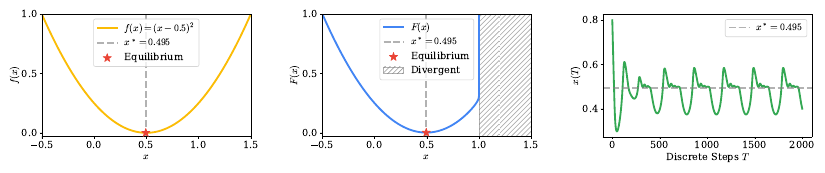}
    \vspace{-1.5em}
    \caption{AdamW with $\eta=10^{-2},\ \beta_1=0.99,\ \beta_2=0.9,\ \lambda=1,\ e=10^{-2}$.  
    Left: quadratic objective $f(x)=(x-\tfrac{1}{2})^2$.  
    Middle: implicit objective $F(x)$ whose stationary point coincides with AdamW’s equilibrium.  
    Right: AdamW iterates $x(T)$ over time, showing non-convergence despite the simple convex objective.}
    \label{fig:threeplots}
\vspace{-0.5em}
\end{figure}

\begin{figure}[t!]
\centering
\includegraphics[width=\textwidth]{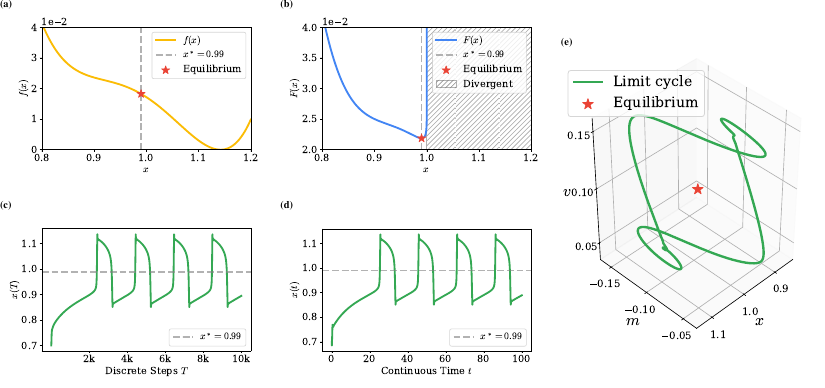}
\vspace{-1.5em}
\caption{AdamW with $\eta=10^{-2},\ \beta_1=0.9,\ \beta_2=0.99,\ \lambda=1,\ e=10^{-3}$.  
(a) Target objective $f(x)$ with fixed point at $x^*=0.99$.  
(b) Implicit objective $F(x)$ whose stationary point coincides with the AdamW fixed point.  
(c) Discrete-time trajectory $x(T)$.  
(d) Continuous-time counterpart $x(t)$.  
(e) Phase portrait in $(x,m,v)$ space. Together, (c–e) show numerical non-convergence and persistent oscillation around the unstable equilibrium.}
\label{fig:fiveplots}
\vspace{-1.5em}
\end{figure}

\paragraph{Implicit objective and stability of AdamW.}
From the fixed-point conditions of AdamW on an objective $f$ with weight decay $\lambda$, we show that its equilibria coincide exactly with the stationary points of a static $\ell_\infty$-constrained and regularized problem, which we call the \emph{implicit objective} $F$ (Theorem~\ref{thm:objective}). This equivalence is only first-order: if AdamW converges, its limit must be a local minimum of $F$, but a local minimum of $F$ need not be a stable equilibrium of AdamW. Stability also requires a curvature condition on $f$: in one dimension, an upper or lower bound on $f''$ depending on the relation between $\beta_1$ and $\beta_2$ (Proposition~\ref{prop:stable_iff}). When $\beta_1>\beta_2$, even simple convex objectives can have unstable fixed points in both continuous and discrete time, independent of the learning rate (\figref{fig:threeplots}). In the common regime $\beta_1<\beta_2$, the obstruction is subtler but still present: non-convex one-dimensional objectives can yield bounded trajectories with persistent oscillations rather than convergence (\figref{fig:fiveplots}). In higher dimensions, a local Lyapunov analysis gives tractable sufficient conditions for stability (\secref{sec:multidimensional}). These behaviors arise from AdamW's update rule, not from adversarial online objectives \citep[cf.][]{reddi2019convergence}.


\paragraph{AdamMe: a convergent modification.}
We trace this instability to a mismatch in AdamW's update: the first-moment estimate $\vm$ is normalized by $\sqrt{\vv}+e$, with second-moment estimate $\vv$ and a small constant $e>0$, whereas the weight decay $\lambda\vx$ on the parameters $\vx$ is not. AdamMe rescales the weight decay by $(\abs{\vm}+e)/(\sqrt{\vv}+e)$, a factor that equals one at every equilibrium, so AdamMe has exactly the same fixed points and implicit objective $F$ as AdamW (Proposition~\ref{prop:adamme_fixed_points}). Unlike AdamW, its continuous-time dynamics admit a Lyapunov function: whenever $1-\beta_2\le 4(1-\beta_1)$ (e.g., $\beta_1=0.9,\ \beta_2=0.999$), every bounded trajectory approaches the equilibrium set (Theorem~\ref{thm:adamme_lyapunov}). The implicit objective is therefore not the source of non-convergence; the dynamics that approach it are. On GPT-2-Small, AdamMe reaches a lower validation loss than AdamW (\figref{fig:adamme_gpt_small}).

\begin{algorithm}[t]
\caption{AdamW and AdamMe. The AdamW update follows the PyTorch implementation \citep{pytorch-adamw}.}
\label{alg:adam_variants}
\begin{algorithmic}[1]
\Require differentiable objective $f:\sR^d\to\sR$
\Require initial parameters $\vx_0$, hyperparameters $\{\eta_t\}_{t=1}^T, \beta_1, \beta_2, e$, weight decay strength $\lambda > 0$
\State Initialize moments $\vm_0 \gets \vzero, \quad \vv_0 \gets \vzero$
\For{$t = 1,2,\dots,T$}
    \State Compute gradient $\vg_t \gets \nabla f(\vx_{t-1})$
    \State Update moment: $\vm_t \gets \beta_1 \vm_{t-1} + (1-\beta_1) \vg_t, \quad \vv_t \gets \beta_2 \vv_{t-1} + (1-\beta_2) \vg_t^{\odot 2}$
    \State Bias correction: $ \hat{\vm}_t \gets \vm_t/(1-\beta_1^t), \quad \hat{\vv}_t \gets \vv_t/(1-\beta_2^t)$
    \If{AdamW}
    \State $\displaystyle \vx_t \gets \vx_{t-1}
	      - \eta_t \Bigl( \frac{\hat{\vm}_t}{\sqrt{\hat{\vv}_t} + e}
	      + \textcolor{blue}{\lambda \vx_{t-1}} \Bigr)$
    \ElsIf{AdamMe}
    \State $\displaystyle \vx_t \gets \vx_{t-1}
	      - \eta_t \Bigl( \frac{\hat{\vm}_t}{\sqrt{\hat{\vv}_t} + e}
	      + \textcolor{blue}{\lambda \frac{|\hat{\vm}_t|+e}{\sqrt{\hat{\vv}_t}+e} \odot \vx_{t-1}} \Bigr)$
    \EndIf
\EndFor
\end{algorithmic}
\end{algorithm}

\section{Background}\label{sec:stabilitybackground}
We analyze AdamW through the ODE obtained in its small-step limit, a standard tool for adaptive optimizers including Adam, Lion, and AdamW \citep{barakat2021convergence, dasilva2020general, malladi2022sdes, chen2023lion, compagnoni2025adaptive}. The limit is well suited to stability analysis: the fixed points of AdamW coincide with the equilibria of its ODE (\secref{sec:nonconverge}), and for small step sizes, their stability carries over to the discrete algorithm \citep{beyn1987numerical, stuart1996dynamical}. The limit thus isolates the update rule itself from step-size effects such as the edge of stability \citep{cohen2021gradient, cohen2022adaptive}, which it does not capture; Appendix~\ref{sec:discrete} treats discrete-time AdamW directly.

We review the classical notions of Lyapunov stability and the \emph{Lyapunov indirect method} (or \emph{linearization method}), which provide standard tools for analyzing the stability of dynamical systems.

\begin{definition}[Lyapunov stability]
Consider the autonomous system $\dot{\vz}(t) = \mT(\vz(t))$, where $\mT:\gD\to\sR^{3d}$ is locally Lipschitz with domain $\gD\subset\sR^{3d}$.  
An equilibrium $\vz_*\in\gD$ is \textbf{Lyapunov stable} if for every $\epsilon>0$ there exists $\delta>0$ such that $\norm{\vz(0)-\vz_*}<\delta$ implies $\norm{\vz(t)-\vz_*}<\epsilon$ for all $t\ge0$. It is \textbf{asymptotically stable} if, additionally, $\vz(t)\to\vz_*$ for all sufficiently close initializations, and \textbf{unstable} if it is not Lyapunov stable.
\end{definition}

Lyapunov's stability establishes the notion of global stability, while the following Hurwitz stability states how the linearized neighborhood near an equilibrium point behaves.
\begin{definition}[Hurwitz stability]
Let $\vz_*$ be an equilibrium of $\dot{\vz} = \mT(\vz)$ and let $\mJ(\vz_*)=\bigl[\partial T_i/\partial z_j\bigr]_{\vz_*}$ denote the Jacobian at $\vz_*$.  
Then $\vz_*$ is \textbf{Hurwitz stable} if every eigenvalue $s$ of $\mJ(\vz_*)$ satisfies $\Re(s)<0$.
\end{definition}

\emph{Lyapunov’s indirect method} gives the following characterization. If $\mJ(\vz_*)$ is Hurwitz stable, then the equilibrium $\vz_*$ is \emph{asymptotically stable}. If $\mJ(\vz_*)$ has an eigenvalue with $\Re(s)>0$, the equilibrium is \emph{unstable}. When all eigenvalues satisfy $\Re(s)\le0$ with some lying on the imaginary axis, the test is inconclusive and higher-order terms must be examined. We direct readers to~\citet{khalil2002nonlinear} for further details.

While Lyapunov’s indirect method addresses continuous-time systems, a parallel criterion applies to discrete dynamics of the form $\vz_{t+1}=\mT(\vz_t)$. In this case, an equilibrium $\vz_*$ is \emph{Schur stable} if all eigenvalues of $\mJ(\vz_*)$ lie strictly inside the unit disk, i.e.\ $|s|<1$ for every eigenvalue $s$. This condition ensures asymptotic stability. If any eigenvalue satisfies $|s|>1$, the equilibrium is unstable. Thus, stability analysis for discrete-time optimization algorithms is analogous to the continuous-time setting.

\paragraph{Convergence.}
Throughout the paper, convergence refers to approach to the optimizer's own stationary state. For a continuous-time system $\dot{\vz}=\mT(\vz)$, this means $\vz(t)\to\vz_*$ for some equilibrium satisfying $\mT(\vz_*)=\vzero$. For a fixed-hyperparameter discrete algorithm $\vz_{t+1}=\mT(\vz_t)$, this means $\vz_t\to\vz_*$ for some fixed point satisfying $\mT(\vz_*)=\vz_*$. When discussing diminishing learning-rate schedules, we do not count convergence caused only by finite total movement; the standard non-vacuous setting requires divergent partial sums, $\sum_t \eta_t=\infty$ (equivalently, $t\to\infty$ in continuous time, where $\sum_t \eta_t$ plays the role of elapsed time), so that convergence reflects vanishing fixed-point residual rather than the iterates merely freezing.

\paragraph{Notation} 
We denote the parameter vector by $\vx$, the first-order momentum by $\vm$, and the second-order momentum by $\vv$. The objective function is $f:\sR^d\to\sR$, defined over the parameter domain. We use $F(\cdot)$ to denote the \emph{implicit objective}, which will be introduced later. Unless stated otherwise, all operations are elementwise. To avoid ambiguity, we use the following terminology: \emph{equilibria} are stationary points of continuous-time dynamical systems, \emph{fixed points} are update-invariant states of discrete-time algorithms, and \emph{stationary points} are points where the gradient of a function vanishes.

\section{Fixed-Point Objective of AdamW}\label{sec:objective}

We start from the \emph{fixed-point conditions} of AdamW, which characterize the relations that convergence points must satisfy if convergence occurs. Decoupled weight decay changes these conditions: AdamW's fixed points no longer align directly with stationary points of the objective $f(\vx)$, but also depend on the weight decay strength.

Formally, let $[\vx_t,\vm_t,\vv_t]$ denote the AdamW iterates. As $t\to\infty$, bias corrections vanish so that $\hat{\vm}_t\to\vm_t$ and $\hat{\vv}_t\to\vv_t$. At a fixed point, the updates are stationary, and the iterates satisfy
\begin{equation}\label{eq:fixpoint}
\left.
\frac{\vm}{\sqrt{\vv}+e}+\lambda\vx=\vzero,\quad
\vm=\nabla f(\vx),\quad
\vv=\nabla f(\vx)^{\odot 2}
\right\}
\Longrightarrow
\frac{\nabla f(\vx)}{\abs{\nabla f(\vx)}+e}+\lambda\vx=\vzero .
\end{equation}
Equation~\eqref{eq:fixpoint} characterizes exactly the fixed points of AdamW. These fixed points turn out to coincide with the stationary points of a constrained and regularized optimization problem, which we refer to as the \emph{implicit objective} of AdamW.

\begin{theorem}\label{thm:objective}
Let $f:\sR^d\to\sR$ be continuously differentiable, and let $\lambda,e>0$. Then $\vx\in\sR^d$ satisfies the fixed-point condition~\eqref{eq:fixpoint} if and only if $\vx$ is a stationary point of the $\ell_\infty$-constrained and regularized problem
\begin{equation}\label{eq:contrained}
    \min_{\vx\in\sR^d}\; F(\vx)=f(\vx)+\tfrac{e}{\lambda}\,\Phi(\lambda\vx),
    \quad \text{subject to}\quad \|\vx\|_\infty\le\tfrac{1}{\lambda},
\end{equation}
where $\Phi(\vx)=\sum_{i=1}^d \phi(x_i)$ with
\[
    \phi(x)=-|x|-\log(1-|x|), \quad |x|<1.
\]
Moreover, since $\phi(x)\to+\infty$ as $|x|\to 1$, the regularizer $\Phi$ serves as a barrier function that naturally enforces the constraint $\|\vx\|_\infty<\tfrac{1}{\lambda}$.
\end{theorem}

\begin{figure}[b]
    \vspace{-1.0em}
    \centering
    \includegraphics[width=0.9\linewidth]{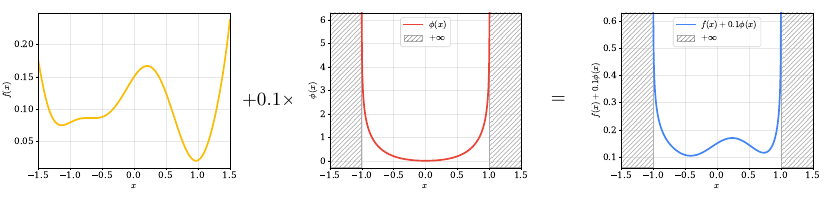}
    \vspace{-1.5em}
    \caption{Left: an objective function $f$ to which AdamW is applied. Middle: $\phi$. Right: $f(x) + \tfrac{e}{\lambda} \phi(\lambda x)$ with $e = 0.1$ and $\lambda = 1$.}
    \label{fig:regularizer}
    \vspace{-1.0em}
\end{figure}

This implicit fixed-point objective coincides with the Softsign case $\nabla \mathcal{K}=x/(|x|+e)$ of Lion-$\mathcal{K}$ (Appendix A, \citet{chen2023lion}), and it reduces to the pure $\ell_\infty$ constraint of \citet{pmlr-v235-xie24e} as $e\to 0$. Although $e$ is typically tiny in practice, $e>0$ is essential for theoretical analysis: without it, convergence to a zero-gradient point of $f$ is not well-defined in AdamW, since the update becomes $0/0$ at any zero-gradient point of $f$, where $m=v=0$. The finite-$e$ formulation is what lets us analyze the stability of such candidate limits. This view also contrasts with prior interpretations of AdamW weight decay \citep{loshchilov2017decoupled, zhou2024towards} as an adaptive $\ell_2$ penalty that depends on optimizer states and therefore departs from a standard static objective. By contrast, our construction yields a static fixed-point objective $F(\vx)$ that depends only on the parameters.

\subsection{First-order coincidence is not stability}
The fact that AdamW’s fixed points coincide with the stationary points of \eqref{eq:contrained} \textbf{does not imply} that AdamW actually \textbf{minimizes} $F(\vx)$. Beyond this first-order consistency, we require: 
\[
\text{\emph{The stable equilibria of AdamW must coincide with the local minima of $F$.}}
\]
For the gradient flow $\dot{\vx}=-\nabla F(\vx)$ this alignment is automatic: $F$ is its own Lyapunov function, so strict local minima are asymptotically stable and nondegenerate saddles are unstable \citep[\S9.3]{hirsch2013differential}, and \citet{lee2016gradient} give the discrete-time analogue for gradient descent. AdamW is not a gradient flow of $F$, so stability is governed by the Jacobian of its own vector field rather than by $\nabla^2F$, and the alignment must be verified separately. Note that the alignment is \textbf{necessary} for AdamW to minimize $F$ but not \textbf{sufficient}: it ensures that any equilibrium reached is a local minimum of $F$, but it does not ensure that trajectories converge at all (\secref{sec:nonconverge}); a global guarantee further requires a Lyapunov function and bounded trajectories, as we establish for AdamMe in Theorem~\ref{thm:adamme_lyapunov}.
For many standard optimizers this alignment is known to hold, including the Lion-$\mathcal{K}$ family \citep{maddison2018hamiltonian, chen2023lion}, whose members include SGD with momentum and Lion \citep{chen2023symbolic}; for Adam without weight decay under a mild condition on $\beta_1,\beta_2$ \citep{barakat2021convergence}; for RMSProp \citep{zaheer2018adaptive} and AdaGrad \citep{defossez2020simple}; and, more recently, for Muon \citep{jordan2024muon, chen2025muon}. AdamW, however, is different: the alignment itself fails. The equivalence between stable equilibria and local minima holds unconditionally only in the degenerate case $\beta_1=\beta_2$. When $\beta_1\ne\beta_2$, as will be shown in the next section, being a local minimum of $F$ is \emph{necessary but not sufficient} for stability. 
\[
\text{\emph{Stable equilibria of AdamW} \ $\subseteq$\ \emph{Local minima of $F$}}
\]
Consequently, some local minima of $F$ are unstable under AdamW. Thus the fixed-point equivalence in Theorem~\ref{thm:objective} is only a first-order characterization, not a descent or convergence guarantee for $F$. It also enables explicit examples where the equilibrium is unstable and numerical trajectories fail to converge to it.

\section{Stability Conditions and Non-Convergence of AdamW}\label{sec:nonconverge}
Throughout this section we consider exact (batch) gradients and the continuous-time limit of infinitesimal step size; unless stated otherwise, stability refers to equilibria of the resulting ODE. This setting is favorable to convergence: it removes gradient noise and learning-rate schedules, and continuous-time dynamics converge under weaker conditions than their discretizations, since gradient flow reaches local minima without smoothness assumptions \citep{Lojasiewicz1984}, while gradient descent also requires smoothness and step-size control \citep{lee2016gradient}. The instabilities we find in this setting therefore reflect the update rule itself.

We model the continuous-time dynamics of AdamW by the ODE
\begin{equation}
    \dot{\vx}_t=-\frac{\vm_t}{\sqrt{\vv_t}+e}-\lambda\vx_t,\quad \dot{\vm}_t=a(\nabla f(\vx_t)-\vm_t),\quad \dot{\vv}_t=b(\nabla f(\vx_t)^{\odot 2}-\vv_t).
\end{equation}
The discrete-time update of AdamW corresponds to a semi-implicit Euler discretization of this system under the identifications $\eta=\Delta t,\ \beta_1=1-a\Delta t,\ \beta_2=1-b\Delta t$. We ignore the bias correction, i.e., we take $\hat{\vm}_t=\vm_t,\ \hat{\vv}_t=\vv_t$, which is exact in the large-$t$ limit.

For the active-coordinate stability calculation below, we set $\vh=\sqrt{\vv}$ and rewrite the system in the equivalent vector form (suppressing $t$ for clarity):
\begin{equation}\label{eq:adamw_ode}
\ddt \vz = \mT(\vz), \quad
\vz \coloneqq \begin{bmatrix}\vx\\\vm\\\vh\end{bmatrix},\quad
\mT(\vz) \coloneqq
\begin{bmatrix}
-\dfrac{\vm}{\vh+e} -\lambda\vx \\[6pt]
a\big(\nabla f(\vx) - \vm\big) \\[6pt]
\dfrac{b}{2}\Big(\dfrac{\nabla f(\vx)^{\odot 2}}{\vh} - \vh\Big)
\end{bmatrix}.
\end{equation}

Let $\vz_*=[\vx_*,\vm_*,\vh_*]$ denote an equilibrium of the dynamics, i.e.\ a point satisfying $\mT(\vz_*)=\vzero$. Differentiating $\mT$ and evaluating at $\vz_*$ gives the Jacobian
\begin{equation}\label{eq:jacobian}
    \mJ(\vz_*)=
    \begin{bmatrix}
    -\lambda\mI & -\diag\!\Big(\dfrac{1}{\vh_*+e}\Big) & \diag\!\Big(\dfrac{\vm_*}{(\vh_*+e)^2}\Big)\\[6pt]
    a\nabla^2 f(\vx_*) & -a\mI & 0\\[6pt]
    b\diag\!\Big(\dfrac{\nabla f(\vx_*)}{\vh_*}\Big)\nabla^2 f(\vx_*) & 0 & -b\mI
    \end{bmatrix}.
\end{equation}

\subsection{Local Stability of AdamW in 1D}
We first analyze the one-dimensional (1D) case, deferring the higher-dimensional analysis to~\secref{sec:multidimensional}. Here ``1D'' refers only to the dimension of $\vx$; the full dynamics remain three-dimensional in $[x,m,h]$. As discussed in~\secref{sec:stabilitybackground}, the local stability of an equilibrium is determined by the spectrum of $\mJ(\vz_*)$. We study this spectrum through the characteristic polynomial $\chi_{\mJ}(s)\coloneqq \det(s\mI-\mJ(\vz_*))$, whose roots are the eigenvalues of $\mJ(\vz_*)$. In the 1D case, this polynomial reduces to a cubic:
\begin{equation}
    \chi_\mJ(s) = s^3 + (a + b + \lambda) s^2 + \left [a b + a \lambda + \frac{(a - b) H_*}{\hbar_*} + \frac{b e \Hat{H}_*}{\hbar_*^2} \right] s + \frac{abe \Hat{H}_*}{\hbar_*^2}. 
\end{equation}
with the shorthand
\[
    \hbar_* \coloneqq h_*+e,
    \quad H_* \coloneqq f''(x_*),
    \quad \Hat{H}_* \coloneqq F''(x_*),
\]
where $F$ is the implicit objective introduced in~\eqref{eq:contrained}. Since $\vz_*$ is an equilibrium, we also use the relations
\[
    h_* = |\nabla f(x_*)| = |m_*|, 
    \quad \Hat{H}_* = H_* + \tfrac{\lambda}{e}\hbar_*^2.
\]

To analyze the signs of the eigenvalues, we recall the following fact.
\begin{lemma}[Routh–Hurwitz criterion for cubics]
\label{lem:rootcubiceq}
All roots of the cubic equation $s^3 + \alpha_2 s^2 + \alpha_1 s + \alpha_0 =0$ have negative real part (i.e., the system is Hurwitz stable) if and only if
\[
\alpha_2,\ \alpha_1,\ \alpha_0 > 0 \quad \text{and} \quad \alpha_2 \alpha_1 > \alpha_0.
\]
\end{lemma}

Lemma~\ref{lem:rootcubiceq} involves two sets of conditions, which we treat in turn. The first set, $\alpha_2,\alpha_1,\alpha_0>0$, shows that every stable equilibrium of AdamW must be a local minimum of the implicit objective $F$ (Corollary~\ref{cor:stable_localmin}). The second set, $\alpha_2\alpha_1>\alpha_0$, shows that the converse fails: local minima of $F$ are stable only under an additional Hessian bound (Proposition~\ref{prop:stable_iff}).

\paragraph{Stable equilibria of AdamW $\subseteq$ Local minima of $F$.}
\begin{corollary}\label{cor:stable_localmin}
Assume $a,b,\lambda,e>0$ and consider the one-dimensional case. If $\vz_*$ is a Hurwitz stable equilibrium of the AdamW ODE, then $x_*$ is a strict local minimum of the implicit objective $F$, i.e.\ $F''(x_*)>0$.
\end{corollary}
\begin{proof}
If $\vz_*$ is Hurwitz stable, then the first set of conditions in Lemma~\ref{lem:rootcubiceq} holds. Since $\alpha_2=a+b+\lambda>0$ always holds and $\alpha_1>0$ follows from $\alpha_0>0$ and $\alpha_2\alpha_1>\alpha_0$, the only informative condition is $\alpha_0=abe\Hat{H}_*/\hbar_*^2>0$, which implies $\Hat{H}_*=F''(x_*)>0$.
\end{proof}

The subtlety, however, is that $F''(x_*)>0$ is necessary but \emph{not sufficient}: even strict local minima of $F$ may correspond to unstable equilibria of AdamW. An additional bound on the Hessian $f''(x_*)$ of the original objective is required.
\begin{proposition}\label{prop:stable_iff}
Assume $a,b,e>0$ and $\lambda\ge0$. Let $[x_*,m_*,h_*]$ be an equilibrium of the AdamW ODE in~\eqref{eq:adamw_ode}. Then $[x_*,m_*,h_*]$ is Hurwitz stable if and only if $F''(x_*)>0$ and 
\begin{equation}\label{eq:hurwitz_condition0}
(a+b+\lambda)\left[ab+a\lambda+\frac{(a-b)f''(x_*)}{\hbar_*}\right] + (b+\lambda)b e \frac{ F''(x_*)}{\hbar_*^2} > 0,
\end{equation}
which can be rearranged into 
\begin{equation}\label{eq:hurwitz_condition}
\left (a (a + \lambda) - b (b + \lambda) \abs{\lambda x_*}\right )  f'' (x_*)  > - \frac{e C}{(1-\abs{\lambda x_*})},
\end{equation}
where $C = (a+b+\lambda)(ab+a\lambda+b\lambda) - ab\lambda > 0$.
\end{proposition}

\begin{wrapfigure}[13]{r}{0.34\textwidth}
    \vspace{-1.1em}
    \centering
    \includegraphics[width=\linewidth]{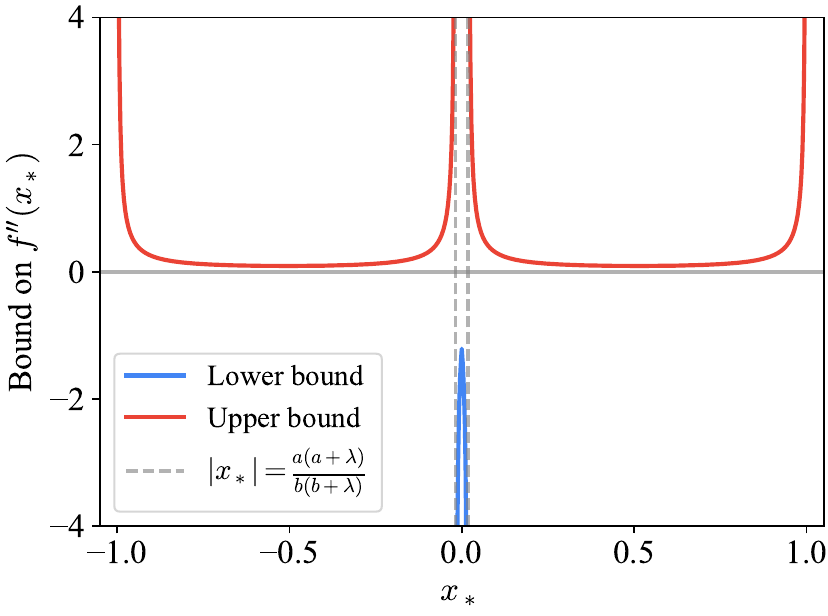}
    \vspace{-1.0em}
    \caption{Hessian constraint in~\eqref{eq:hurwitz_condition} for $a=1,\ b=10$.}
    \label{fig:bound}
    \vspace{-0.6em}
\end{wrapfigure}

Stability therefore requires two conditions: the implicit Hessian must satisfy $F''(x_*)>0$, and the inequality~\eqref{eq:hurwitz_condition} must hold, which imposes an additional constraint on the Hessian $f''(x_*)$ of the original loss $f$. We illustrate this constraint in~\figref{fig:bound} for the regime $a<b$: the sign of the coefficient multiplying $f''(x_*)$ changes with $\abs{x_*}$, so the condition can switch between a lower and an upper bound. In this sense, one may say
\[
  \textit{AdamW is a faithful but selective optimizer of $F(x)$.}
\]
Equivalently, AdamW can be interpreted as minimizing $F(x)$, but only under the additional constraint imposed by~\eqref{eq:hurwitz_condition}.

\begin{remark}[When the additional condition vanishes]
The extra requirement in~\eqref{eq:hurwitz_condition} disappears in two important cases:
\begin{enumerate}
    \item \textbf{Adam limit ($\lambda \to 0$).} The regularizer and constraint vanish, so the implicit objective reduces to $F(x)=f(x)$. The extra requirement becomes $a^2 f''(x_*)>-eC$, which is automatically implied by $F''(x_*)=f''(x_*)>0$. Stability then reduces to the standard requirement that $x_*$ be a strict local minimum of $f$, showing that the additional complication arises entirely from the weight-decay term.
    \item \textbf{Equal momentum coefficients ($a=b$).}
    In this case,~\eqref{eq:hurwitz_condition0} simplifies to $F''(x_*)>-L$ for some $L\geq0$, which is implied by $F''(x_*)>0$. Hence stability holds if and only if $x_*$ is a strict local minimum of $F$ (see Appendix~\ref{cor:vanish}).
\end{enumerate}
In both cases stability reduces to $F''(x_*)>0$, so the extra bound arises only when weight decay is combined with unequal momentum coefficients $a$ and $b$.
\end{remark}

We present the analysis in continuous time because the limit isolates the mechanism specific to AdamW. The exact discrete-time conditions (Appendix~\ref{sec:discrete}) contain the same requirement $F''(x_*)>0$ and a version of~\eqref{eq:hurwitz_condition} whose coefficients depend on $\eta$, together with the familiar large-step-size ceiling, the analogue of $\eta<2/L$ for gradient descent. The limit removes this ceiling and the $\eta$-dependence, leaving~\eqref{eq:hurwitz_condition} in closed form.

\subsection{Examples of Non-Convergence}
Whenever $a\neq b$, one can construct objectives $f$ that violate the stability condition~\eqref{eq:hurwitz_condition0} while still satisfying $F''(x_*)>0$. Such objectives yield strict local minima of $F$, the very points an optimizer is meant to find, that AdamW's own dynamics fail to stabilize. The type of violation depends on the relation between $a$ and $b$. When $a<b$ (equivalently, $\beta_2<\beta_1$), the condition fails for sufficiently large $f''$, so even strongly convex functions can break stability — a step-size-independent obstruction in a regime already known to perform poorly in practice. When $a>b$ (equivalently, $\beta_2>\beta_1$), which is the regime most common in practice, instability arises only for non-convex objectives and must be constructed more carefully. We now provide simple examples illustrating both cases; in the common regime, the simulations exhibit bounded trajectories with persistent oscillations rather than convergence to the unstable equilibrium.

\begin{example}
When $\beta_2 < \beta_1$ (so $a<b$),~\eqref{eq:hurwitz_condition} imposes an \emph{upper} bound $f''(x_*)\le U$, which can fail even for simple convex functions. Consider the quadratic
$$
f(x)=(x-0.5)^2.
$$
With $\lambda=1$ and $e=10^{-2}$, the equilibrium is $x_*\approx0.49$. Choosing $\beta_1=0.99,\ \beta_2=0.9$ and step size $\eta=10^{-2}$ yields $a=1,\ b=10$, so~\eqref{eq:hurwitz_condition} requires $f''(x_*)<9.1\times10^{-2}$; but $f''=2$, hence the equilibrium is unstable. The corresponding discrete fixed point is also unstable at $\eta=10^{-2}$, as shown in~\figref{fig:threeplots}.
\end{example}

\begin{example}
When $\beta_2 > \beta_1$ (so $a>b$), the condition imposes a \emph{lower} bound $f''(x_*)\ge -L$, which can be violated by a carefully chosen non-convex quartic. Consider
$$
f(x)=20(x-0.99)^4-0.60(x-0.99)^2-0.099(x-0.99)+0.018.
$$
With $a=10,\ b=1$ (corresponding to $\beta_1=0.90,\ \beta_2=0.99$), $\eta=10^{-2}$, $e=10^{-3}$, and $\lambda=1$, the equilibrium at $x_*=0.99$ has $f''(x_*)\approx-1.20$, violating the required bound $f''(x_*)>-0.22$. The equilibrium is therefore unstable, and the displayed AdamW trajectory again fails to converge to it (see~\figref{fig:fiveplots}).
\end{example}

We construct these worst-case counterexamples on controlled objectives to show that AdamW admits no general convergence guarantee without additional assumptions. While our analysis yields theoretical insight into AdamW's dynamics, its practical implications for model training are more subtle: modern training is evaluated by validation performance rather than asymptotic parameter convergence, and LLMs routinely achieve strong downstream performance despite non-convergent or non-monotone training dynamics \citep{chandramoorthy2022generalization, cohen2021gradient}.

\subsection{A Convergent Modification}\label{sec:adamme}

The instability above suggests modifying the weight-decay term while preserving AdamW's fixed-point equation. We replace the fixed weight decay $\lambda\vx_t$ with an adaptive one that scales with the strength of AdamW's gradient update, yielding the ODE dynamics
\begin{equation}\label{eq:adamme_ode}
    \dot{\vx}_t=-\frac{\vm_t}{\sqrt{\vv_t}+e}-\lambda \frac{\abs{\vm_t} + e}{\sqrt{\vv_t}+e} \vx_t,\quad
    \dot{\vm}_t=a(\nabla f(\vx_t)-\vm_t),\quad
    \dot{\vv}_t=b(\nabla f(\vx_t)^{\odot 2}-\vv_t).
\end{equation}
We call this modification AdamMe. At equilibrium, $\abs{\vm_*}=\sqrt{\vv_*}$, so the additional multiplier equals one and AdamW's fixed-point structure is preserved. The corresponding discrete update is given in Algorithm~\ref{alg:adam_variants}. AdamMe is designed to change only how the dynamics move, not where they come to rest: as shown in the following proposition, it targets exactly the same solutions as AdamW.

\begin{proposition}\label{prop:adamme_fixed_points}
The equilibria of~\eqref{eq:adamme_ode} coincide with the fixed points of AdamW. Consequently, AdamMe has the same implicit fixed-point objective $F$ as in Theorem~\ref{thm:objective}.
\end{proposition}

\begin{proof}
At equilibrium, $\vm_*=\nabla f(\vx_*)$ and $\sqrt{\vv_*}=\abs{\nabla f(\vx_*)}=\abs{\vm_*}$. Substituting these identities into the $\vx$-equation of~\eqref{eq:adamme_ode} gives
\[
    \frac{\nabla f(\vx_*)}{\abs{\nabla f(\vx_*)}+e}+\lambda \vx_*=\vzero,
\]
which is exactly the AdamW fixed-point condition~\eqref{eq:fixpoint}. The equivalence with the stationary points of $F$ then follows from Theorem~\ref{thm:objective}.
\end{proof}

\begin{theorem}[Lyapunov convergence of AdamMe]\label{thm:adamme_lyapunov}
Assume $f$ is $C^1$ with locally Lipschitz gradient, $a,b,\lambda,e>0$, and $b\le 4a$. Along every solution of~\eqref{eq:adamme_ode} initialized with $\lambda |x_i(0)|<1$ and $v_i(0)>0$, this domain is forward invariant and the Lyapunov function $V$ defined in Appendix~\ref{sec:adamme_lyapunov} is nonincreasing. Consequently, every bounded trajectory approaches the equilibrium set; if the limiting equilibria are isolated, the trajectory converges to one equilibrium.
\end{theorem}

\begin{wrapfigure}[14]{r}{0.4\textwidth}
    \vspace{-1.5em}
    \centering
    \includegraphics[width=\linewidth]{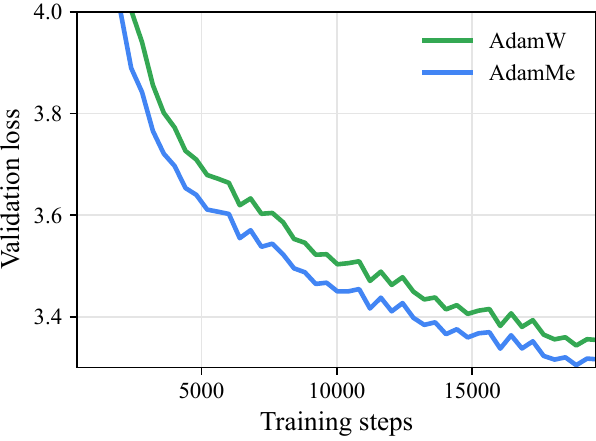}
    \vspace{-1.5em}
    \caption{GPT-2-Small validation loss. AdamMe tracks AdamW and achieves better validation loss (3.317 vs.\ 3.355).}
    \label{fig:adamme_gpt_small}
    \vspace{-2.5em}
\end{wrapfigure}

The explicit Lyapunov function and proof are given in Appendix~\ref{sec:adamme_lyapunov}. Thus AdamMe preserves the same implicit objective as AdamW while restoring a convergence guarantee for the modified continuous-time dynamics. This is a constructive counterpart to the negative result: the implicit objective itself is not the source of non-convergence; rather, the issue lies in how AdamW's dynamics approach its fixed points.

As a sanity check, we evaluate AdamMe on a small language-model run following~\citet{amsel2025polar} with its GPT-2-Small (124M) configuration. As shown in~\figref{fig:adamme_gpt_small}, AdamMe closely tracks AdamW and reaches a slightly lower validation loss under hyperparameters tuned for each method, suggesting that the correction preserves AdamW-like training behavior with a small performance gain.

\section{Multi-Dimensional Stability of AdamW}
\label{sec:multidimensional}

In the multi-dimensional setting, the Hessian and AdamW's coordinatewise preconditioner interact as matrices, so the scalar cubic analysis no longer gives a compact stability boundary. While local stability can be checked numerically from the full Jacobian spectrum, this gives little theoretical insight into when stability holds. We therefore give a minimal sufficient condition using a Lyapunov function. The linearized dynamics are given by the following lemma.
\begin{lemma}\label{lem:linear_ode}
Let $\vz_* = (\vx_*, \vm_*, \vh_*)$ be an equilibrium point of the AdamW ODE~\eqref{eq:adamw_ode}.  
Then the linearized dynamics $\dot{\vz}_t = \mJ(\vz_*) \vz_t$ imply that $\vx_t$ satisfies
\begin{equation}\label{eq:linear_ode}
    \dddot{\vx}_t 
    + \mA_2 \ddot{\vx}_t
    + \mA_1 \dot{\vx}_t
    + \mA_0 \vx_t = 0,
\end{equation}
where
\begin{align*}
    \mA_2 &= (a+b+\lambda) \mI \quad \mA_0 = ab\left[\,\diag(e/(\vh_* + e)^2) \nabla^2 f(\vx_*) + \lambda \mI\,\right] \\
    \mA_1 &= (ab + a\lambda + b\lambda)\mI
            + (a-b)\,\diag(1/(\vh_* + e)) \nabla^2 f(\vx_*)
            + b\,\diag(e/(\vh_* + e)^2) \nabla^2 f(\vx_*), \\
\end{align*}
\end{lemma}
\vspace{-0.5em}
We can further rewrite~\eqref{eq:linear_ode} as a $3d$-dimensional first-order ODE system:  
\[
\dot\vx_t=\vy_t,\quad 
\dot\vy_t=\vz_t,\quad 
\dot\vz_t=-\mA_0\vx_t-\mA_1\vy_t-\mA_2\vz_t.
\]
To analyze stability, we construct a Lyapunov function of the form 
\[
V(\vx,\vy,\vz)=\tfrac{1}{2}\Bigl(\|\mA_2\vx+\vy\|_{\mP}^2+\|\mA_2\vy+\vz\|_{\mQ}^2+\|\vy\|_{\mK}^2\Bigr),
\]
where $\mP,\mQ,\mK$ are defined as $\mP=\mQ\mA_0\mA_2^{-1}$, $\mK=\mQ(\mA_1-\mA_0\mA_2^{-1})$, and $\mL=\mA_2^\top\mQ(\mA_1-\mA_0\mA_2^{-1})$, with $\mQ$ chosen so that $\mP,\mQ,\mK$ are positive definite. With these definitions, we obtain
\[
\ddt V(\vx_t,\vy_t,\vz_t)=-\tfrac{1}{2}\vy^\top(\mL+\mL^\top)\vy.
\]
Thus $\dot V \leq 0$ whenever $\mL+\mL^\top$ is positive definite, leading to the following result.

\begin{proposition}\label{prop:multidim_stability}
Consider the system in~\eqref{eq:linear_ode}.  
The dynamics are asymptotically stable if there exists a symmetric positive definite matrix $\mQ$ such that 1.~$\mQ\mA_0$ and $\mQ\mA_1$ are symmetric, and 2.~$\mQ\mA_0$, $\mQ\mA_1$, and $\mQ(\mA_1\mA_2-\mA_0)$ are positive definite.

\end{proposition}

Here $\mQ$ specifies the metric used in the Lyapunov function. Once an equilibrium is fixed, $\mA_0,\mA_1,\mA_2$ are fixed by the local Hessian and AdamW preconditioner, and the proposition asks whether some $\mQ$ satisfies the stated symmetry and positivity conditions. This constitutes a minimal sufficient condition and exact local stability is determined by the spectrum of the full $3d\times3d$ Jacobian.

\section{Related Works}\label{sec:related_works}

\paragraph{Convergence analysis of Adam.} 
\citet{kingma2014adam} introduced Adam with a convergence proof for online convex optimization, which \citet{reddi2019convergence} later showed to contain a gap. Their counterexample, however, relies on carefully constructed adversarial online objectives and does not preclude convergence for a fixed problem, where Adam converges once $\beta_2$ is large enough and $\beta_1<\sqrt{\beta_2}$ \citep{zhang2022adam, zhang2026adam}. Follow-up work proved convergence for non-convex functions under bounded gradients and consistent gradient signs \citep{de2018convergence, defossez2020simple}, later relaxed to unbounded gradients \citep{zhang2022adam} and to non-uniform smoothness and affine-variance noise \citep{wang2024provable, li2023convergence, hong2024convergence, jin2024comprehensive}, narrowing the gap with SGD; \citet{dereich2025sharp} further derived sharp local rates near strict minimizers. On the negative side, \citet{dereich2025symmetry} showed that even on simple stochastic quadratics, Adam converges to the minimizer only when the data are symmetrically distributed. A complementary line of work studies Adam's implicit bias to explain its empirical success \citep{xie2024adam, vasudeva2025rich, li2025adam}.

\paragraph{Adam with decoupled weight decay.} 
In practice, AdamW \citep{loshchilov2017decoupled}, the decoupled weight decay variant of Adam, has become the default choice over the original Adam. Weight decay itself was introduced by \citet{hanson1988comparing}, and while it is equivalent to $L_2$ regularization in SGD, \citet{loshchilov2017decoupled} showed this equivalence breaks for adaptive methods and that decoupling improves performance. Since then, the effectiveness of weight decay has been confirmed across architectures and scales \citep{kosson2023rotational, d2024we, wang2024set, bergsma2025power, he2025alphadecay, chen2025cautious}.

\paragraph{Continuous-time optimizer analysis.}
Continuous-time analysis has long been used to study optimizer stability and acceleration: heavy-ball dynamics give a dissipative ODE model \citep{attouch2000heavy}, the Nesterov ODE explains acceleration and restart behavior \citep{su2016differential}, and later work extends this viewpoint to accelerated mirror descent, Bregman/Hamiltonian flows, Lyapunov analyses, and numerical discretizations \citep{krichene2015accelerated, wibisono2016variational, scieur2017integration, shi2022understanding, muehlebach2021optimization, liang2024memory, liang2024cautious, nguyen2024memory, chen2025muon}. For stochastic algorithms, modified-equation and diffusion-limit analyses approximate SGD and adaptive methods in continuous time \citep{li2017stochastic, li2019stochastic, orvieto2019continuous, malladi2022sdes}, while \citet{barakat2021convergence} develop an Adam-specific ODE and convergence analysis. Our use of an AdamW ODE is different in focus: we characterize the static fixed-point objective induced by decoupled weight decay, then show that its stationary points need not be stable equilibria of the AdamW dynamics.

\paragraph{No conflicts with previous studies of AdamW's convergence.}
Several works have analyzed AdamW’s convergence, but under assumptions that depart from the algorithm as commonly used. \citet{zhou2024towards} considered exponentially decaying weight decay $\lambda_k=\lambda(1-\alpha)^k$, in contrast to the standard constant choice, and defined convergence through $\frac{1}{T}\sum_{k=0}^{T-1}\E[\|\nabla F_k(\vx_k)\|_2^2]$, where $F_k$ is a state-dependent ``dynamically regularized'' objective. Recently \citet{li2025frac} established a rate of $\mathcal{O}(d^{1/2}T^{-1/4})$ for the criterion $\frac{1}{T}\sum_{k=0}^{T-1}\E[\|\nabla f(\vx_k)\|_1]\le\epsilon^2$, but their result requires that the iterates already satisfy the $\ell_\infty$ constraint and assumes small weight decay ($\lambda\le\mathcal{O}(d^{1/2}T^{-3/4})$) together with small momentum coefficients ($\beta_1=\mathcal{O}(T^{-1/2})$). \citet{pmlr-v235-xie24e} examined AdamW’s implicit objective in the limit $e\to0$, showing convergence to KKT points under suitable learning-rate conditions, but only conditional on AdamW itself converging. In contrast, our results demonstrate that AdamW fails to converge under standard hyperparameters without additional assumptions, and thus do not conflict with previous analyses.

\section{Conclusions}
\label{sec:conclusions}
We studied the dynamics of AdamW and showed that, despite admitting an implicit objective, its equilibria can be unstable; on simple problems, numerical trajectories then exhibit persistent oscillations rather than convergence. Extending the analysis to higher dimensions, we derived a sufficient condition for local stability via a Lyapunov function. As a constructive byproduct, an adaptive rescaling of the weight decay (AdamMe) preserves AdamW's fixed points and admits a convergence guarantee. Together, these results reveal that AdamW's theoretical properties are unexpectedly fragile: its stability depends on delicate interactions between curvature, weight decay, and momentum, yet the method remains the default optimizer in practice. This contrast points to room for principled refinements that retain AdamW's empirical strengths while offering stronger guarantees.

\paragraph{Limitations.}
First, in higher dimensions we give only sufficient conditions for the stability of AdamW (\secref{sec:multidimensional}). For any given configuration, the eigenvalues of the $3d\times3d$ Jacobian~\eqref{eq:jacobian} can still be computed with computer-aided tools to decide Hurwitz stability exactly, but the resulting closed-form expressions are too unwieldy to read, so we present compact sufficient conditions instead. Second, although we show that AdamW fails to converge on some objectives, how severe this is for large-scale neural network training remains unclear. Neural networks need not converge to be useful, and in our experiments training appears robust to this instability; it has not prevented the success of the many foundation models trained with AdamW \citep{grattafiori2024llama, liu2024deepseek}. Third, the stability conditions for discrete-time AdamW are complicated even in one dimension. We therefore report these conditions, together with visualizations, in Appendix~\ref{sec:discrete}.

\begin{ack}
This work was supported in part by the Institute for Foundations of Machine Learning (IFML).
The authors acknowledge the Texas Advanced Computing Center (TACC) at The University of Texas at
Austin for providing computational resources that have contributed to the research results reported
within this paper.
\end{ack}

\clearpage
\bibliography{references}

@article{kingma2014adam,
  title={Adam: A method for stochastic optimization},
  author={Kingma, Diederik P and Ba, Jimmy},
  journal={arXiv preprint arXiv:1412.6980},
  year={2014}
}

@article{grattafiori2024llama,
  title={The llama 3 herd of models},
  author={Grattafiori, Aaron and Dubey, Abhimanyu and Jauhri, Abhinav and Pandey, Abhinav and Kadian, Abhishek and Al-Dahle, Ahmad and Letman, Aiesha and Mathur, Akhil and Schelten, Alan and Vaughan, Alex and others},
  journal={arXiv preprint arXiv:2407.21783},
  year={2024}
}

@article{liu2024deepseek,
  title={Deepseek-v3 technical report},
  author={Liu, Aixin and Feng, Bei and Xue, Bing and Wang, Bingxuan and Wu, Bochao and Lu, Chengda and Zhao, Chenggang and Deng, Chengqi and Zhang, Chenyu and Ruan, Chong and others},
  journal={arXiv preprint arXiv:2412.19437},
  year={2024}
}

@inproceedings{esser2024scaling,
  title={Scaling rectified flow transformers for high-resolution image synthesis},
  author={Esser, Patrick and Kulal, Sumith and Blattmann, Andreas and Entezari, Rahim and M{\"u}ller, Jonas and Saini, Harry and Levi, Yam and Lorenz, Dominik and Sauer, Axel and Boesel, Frederic and others},
  booktitle={Forty-first international conference on machine learning},
  year={2024}
}

@article{ravi2024sam,
  title={Sam 2: Segment anything in images and videos},
  author={Ravi, Nikhila and Gabeur, Valentin and Hu, Yuan-Ting and Hu, Ronghang and Ryali, Chaitanya and Ma, Tengyu and Khedr, Haitham and R{\"a}dle, Roman and Rolland, Chloe and Gustafson, Laura and others},
  journal={arXiv preprint arXiv:2408.00714},
  year={2024}
}

@article{loshchilov2017decoupled,
  title={Decoupled weight decay regularization},
  author={Loshchilov, Ilya and Hutter, Frank},
  journal={arXiv preprint arXiv:1711.05101},
  year={2017}
}

@article{reddi2019convergence,
  title={On the convergence of adam and beyond},
  author={Reddi, Sashank J and Kale, Satyen and Kumar, Sanjiv},
  journal={arXiv preprint arXiv:1904.09237},
  year={2019}
}

@article{zhang2022adam,
  title={Adam can converge without any modification on update rules},
  author={Zhang, Yushun and Chen, Congliang and Shi, Naichen and Sun, Ruoyu and Luo, Zhi-Quan},
  journal={Advances in neural information processing systems},
  volume={35},
  pages={28386--28399},
  year={2022}
}

@article{de2018convergence,
  title={Convergence guarantees for RMSProp and ADAM in non-convex optimization and an empirical comparison to Nesterov acceleration},
  author={De, Soham and Mukherjee, Anirbit and Ullah, Enayat},
  journal={arXiv preprint arXiv:1807.06766},
  year={2018}
}

@article{defossez2020simple,
  title={A simple convergence proof of adam and adagrad},
  author={D{\'e}fossez, Alexandre and Bottou, L{\'e}on and Bach, Francis and Usunier, Nicolas},
  journal={arXiv preprint arXiv:2003.02395},
  year={2020}
}

@article{li2023convergence,
  title={Convergence of adam under relaxed assumptions},
  author={Li, Haochuan and Rakhlin, Alexander and Jadbabaie, Ali},
  journal={Advances in Neural Information Processing Systems},
  volume={36},
  pages={52166--52196},
  year={2023}
}

@article{hanson1988comparing,
  title={Comparing biases for minimal network construction with back-propagation},
  author={Hanson, Stephen and Pratt, Lorien},
  journal={Advances in neural information processing systems},
  volume={1},
  year={1988}
}

@article{kosson2023rotational,
  title={Rotational equilibrium: How weight decay balances learning across neural networks},
  author={Kosson, Atli and Messmer, Bettina and Jaggi, Martin},
  journal={arXiv preprint arXiv:2305.17212},
  year={2023}
}

@article{wang2024set,
  title={How to set AdamW's weight decay as you scale model and dataset size},
  author={Wang, Xi and Aitchison, Laurence},
  journal={arXiv preprint arXiv:2405.13698},
  year={2024}
}

@InProceedings{pmlr-v235-xie24e,
  title = 	 {Implicit Bias of {A}dam{W}: $\ell_\infty$-Norm Constrained Optimization},
  author =       {Xie, Shuo and Li, Zhiyuan},
  booktitle = 	 {Proceedings of the 41st International Conference on Machine Learning},
  pages = 	 {54488--54510},
  year = 	 {2024},
  editor = 	 {Salakhutdinov, Ruslan and Kolter, Zico and Heller, Katherine and Weller, Adrian and Oliver, Nuria and Scarlett, Jonathan and Berkenkamp, Felix},
  volume = 	 {235},
  series = 	 {Proceedings of Machine Learning Research},
  month = 	 {21--27 Jul},
  publisher =    {PMLR},
  url = 	 {https://proceedings.mlr.press/v235/xie24e.html}
}

@article{d2024we,
  title={Why do we need weight decay in modern deep learning?},
  author={D'Angelo, Francesco and Andriushchenko, Maksym and Varre, Aditya Vardhan and Flammarion, Nicolas},
  journal={Advances in Neural Information Processing Systems},
  volume={37},
  pages={23191--23223},
  year={2024}
}

@inproceedings{Lojasiewicz1984,
  author    = {Stanisław {\L}ojasiewicz},
  title     = {Sur les trajectoires du gradient d’une fonction analytique},
  booktitle = {Séminari di Geometria 1982--1983},
  year      = {1984},
  pages     = {115--117},
  address   = {Bologna, Italy},
  publisher = {Istituto di Geometria, Dipartimento di Matematica, Università di Bologna}
}

@inproceedings{lee2016gradient,
  title={Gradient descent only converges to minimizers},
  author={Lee, Jason D and Simchowitz, Max and Jordan, Michael I and Recht, Benjamin},
  booktitle={Conference on learning theory},
  pages={1246--1257},
  year={2016},
  organization={PMLR}
}

@article{zhou2024towards,
  title={Towards understanding convergence and generalization of AdamW},
  author={Zhou, Pan and Xie, Xingyu and Lin, Zhouchen and Yan, Shuicheng},
  journal={IEEE transactions on pattern analysis and machine intelligence},
  volume={46},
  number={9},
  pages={6486--6493},
  year={2024},
  publisher={IEEE}
}

@article{li2025frac,
  title = {On the $O\!\left(\frac{\sqrt{d}}{K^{1/4}}\right)$ Convergence Rate of AdamW Measured by $\ell_1$ Norm},
  author = {Li, Huan and Dong, Yiming and Lin, Zhouchen},
  journal = {arXiv preprint arXiv:2505.11840},
  year = {2025}
}

@manual{pytorch-adamw,
  title        = {torch.optim.AdamW — PyTorch Documentation},
  author       = {{PyTorch Contributors}},
  organization = {PyTorch},
  version      = {2.8},
  year         = {2025},
  note         = {Accessed: September 19, 2025. \url{https://docs.pytorch.org/docs/stable/generated/torch.optim.AdamW.html}},
}

@article{chen2023lion,
  title={Lion secretly solves constrained optimization: As lyapunov predicts},
  author={Chen, Lizhang and Liu, Bo and Liang, Kaizhao and Liu, Qiang},
  journal={arXiv preprint arXiv:2310.05898},
  year={2023},
  note={Version 7, revised July 2025},
  url={https://arxiv.org/abs/2310.05898v7}
}

@article{chen2023symbolic,
  title={Symbolic discovery of optimization algorithms},
  author={Chen, Xiangning and Liang, Chen and Huang, Da and Real, Esteban and Wang, Kaiyuan and Pham, Hieu and Dong, Xuanyi and Luong, Thang and Hsieh, Cho-Jui and Lu, Yifeng and others},
  journal={Advances in neural information processing systems},
  volume={36},
  pages={49205--49233},
  year={2023}
}

@misc{jordan2024muon,
  author       = {Keller Jordan and Yuchen Jin and Vlado Boza and You Jiacheng and
                  Franz Cesista and Laker Newhouse and Jeremy Bernstein},
  title        = {Muon: An optimizer for hidden layers in neural networks},
  year         = {2024},
  url          = {https://kellerjordan.github.io/posts/muon/}
}

@article{maddison2018hamiltonian,
  title={Hamiltonian descent methods},
  author={Maddison, Chris J and Paulin, Daniel and Teh, Yee Whye and O'Donoghue, Brendan and Doucet, Arnaud},
  journal={arXiv preprint arXiv:1809.05042},
  year={2018}
}

@article{vasudeva2025rich,
  title={The Rich and the Simple: On the Implicit Bias of Adam and SGD},
  author={Vasudeva, Bhavya and Lee, Jung Whan and Sharan, Vatsal and Soltanolkotabi, Mahdi},
  journal={arXiv preprint arXiv:2505.24022},
  year={2025}
}

@article{li2025adam,
  title={Adam Reduces a Unique Form of Sharpness: Theoretical Insights Near the Minimizer Manifold},
  author={Li, Xinghan and Wen, Haodong and Lyu, Kaifeng},
  journal={arXiv preprint arXiv:2511.02773},
  year={2025}
}

@article{jin2024comprehensive,
  title={A Comprehensive Framework for Analyzing the Convergence of Adam: Bridging the Gap with SGD},
  author={Jin, Ruinan and Li, Xiao and Yu, Yaoliang and Wang, Baoxiang},
  journal={arXiv preprint arXiv:2410.04458},
  year={2024}
}

@article{xie2024adam,
  title={Adam Exploits $\ell_\infty$ geometry of Loss Landscape via Coordinate-wise Adaptivity},
  author={Xie, Shuo and Mohamadi, Mohamad Amin and Li, Zhiyuan},
  journal={arXiv preprint arXiv:2410.08198},
  year={2024}
}

@article{bergsma2025power,
  title={Power lines: Scaling laws for weight decay and batch size in llm pre-training},
  author={Bergsma, Shane and Dey, Nolan and Gosal, Gurpreet and Gray, Gavia and Soboleva, Daria and Hestness, Joel},
  journal={arXiv preprint arXiv:2505.13738},
  year={2025}
}

@article{he2025alphadecay,
  title={AlphaDecay: Module-wise Weight Decay for Heavy-Tailed Balancing in LLMs},
  author={He, Di and Jaiswal, Ajay and Tu, Songjun and Shen, Li and Yuan, Ganzhao and Liu, Shiwei and Yin, Lu},
  journal={arXiv preprint arXiv:2506.14562},
  year={2025}
}

@book{khalil2002nonlinear,
  author    = {Khalil, Hassan K.},
  title     = {Nonlinear Systems},
  edition   = {3rd},
  year      = {2002},
  publisher = {Prentice Hall},
  address   = {Upper Saddle River, NJ},
  isbn      = {0130673897},
  pages     = {750}
}

@article{attouch2000heavy,
  title={The heavy ball with friction method, I. The continuous dynamical system: Global exploration of the local minima of a real-valued function by asymptotic analysis of a dissipative dynamical system},
  author={Attouch, Hedy and Goudou, Xavier and Redont, Patrick},
  journal={Communications in Contemporary Mathematics},
  volume={2},
  number={1},
  pages={1--34},
  year={2000},
  doi={10.1142/S0219199700000025}
}

@article{su2016differential,
  title={A Differential Equation for Modeling Nesterov's Accelerated Gradient Method: Theory and Insights},
  author={Su, Weijie and Boyd, Stephen and Cand{\`e}s, Emmanuel J.},
  journal={Journal of Machine Learning Research},
  volume={17},
  number={153},
  pages={1--43},
  year={2016}
}

@inproceedings{krichene2015accelerated,
  title={Accelerated Mirror Descent in Continuous and Discrete Time},
  author={Krichene, Walid and Bayen, Alexandre and Bartlett, Peter L.},
  booktitle={Advances in Neural Information Processing Systems},
  volume={28},
  year={2015}
}

@article{wibisono2016variational,
  title={A Variational Perspective on Accelerated Methods in Optimization},
  author={Wibisono, Andre and Wilson, Ashia C. and Jordan, Michael I.},
  journal={Proceedings of the National Academy of Sciences},
  volume={113},
  number={47},
  pages={E7351--E7358},
  year={2016},
  doi={10.1073/pnas.1614734113}
}

@inproceedings{scieur2017integration,
  title={Integration Methods and Optimization Algorithms},
  author={Scieur, Damien and Roulet, Vincent and Bach, Francis and d'Aspremont, Alexandre},
  booktitle={Advances in Neural Information Processing Systems},
  volume={30},
  year={2017}
}

@article{shi2022understanding,
  title={Understanding the Acceleration Phenomenon via High-Resolution Differential Equations},
  author={Shi, Bin and Du, Simon S. and Jordan, Michael I. and Su, Weijie J.},
  journal={Mathematical Programming},
  volume={195},
  pages={79--148},
  year={2022},
  doi={10.1007/s10107-021-01681-8}
}

@article{muehlebach2021optimization,
  title={Optimization with Momentum: Dynamical, Control-Theoretic, and Symplectic Perspectives},
  author={Muehlebach, Michael and Jordan, Michael I.},
  journal={Journal of Machine Learning Research},
  volume={22},
  number={73},
  pages={1--50},
  year={2021}
}

@inproceedings{li2017stochastic,
  title={Stochastic Modified Equations and Adaptive Stochastic Gradient Algorithms},
  author={Li, Qianxiao and Tai, Cheng and E, Weinan},
  booktitle={Proceedings of the 34th International Conference on Machine Learning},
  pages={2101--2110},
  year={2017},
  volume={70},
  series={Proceedings of Machine Learning Research},
  publisher={PMLR}
}

@article{li2019stochastic,
  title={Stochastic Modified Equations and Dynamics of Stochastic Gradient Algorithms I: Mathematical Foundations},
  author={Li, Qianxiao and Tai, Cheng and E, Weinan},
  journal={Journal of Machine Learning Research},
  volume={20},
  number={40},
  pages={1--47},
  year={2019}
}

@inproceedings{orvieto2019continuous,
  title={Continuous-Time Models for Stochastic Optimization Algorithms},
  author={Orvieto, Antonio and Lucchi, Aurelien},
  booktitle={Advances in Neural Information Processing Systems},
  volume={32},
  year={2019}
}

@inproceedings{malladi2022sdes,
  title={On the SDEs and Scaling Rules for Adaptive Gradient Algorithms},
  author={Malladi, Sadhika and Lyu, Kaifeng and Panigrahi, Abhishek and Arora, Sanjeev},
  booktitle={Advances in Neural Information Processing Systems},
  volume={35},
  year={2022}
}

@article{barakat2021convergence,
  title={Convergence and Dynamical Behavior of the ADAM Algorithm for Nonconvex Stochastic Optimization},
  author={Barakat, Anas and Bianchi, Pascal},
  journal={SIAM Journal on Optimization},
  volume={31},
  number={1},
  pages={244--274},
  year={2021},
  doi={10.1137/19M1263443}
}

@inproceedings{chandramoorthy2022generalization,
  title={On the Generalization of Learning Algorithms That Do Not Converge},
  author={Chandramoorthy, Nisha and Loukas, Andreas and Gatmiry, Khashayar and Jegelka, Stefanie},
  booktitle={Advances in Neural Information Processing Systems},
  volume={35},
  pages={34241--34257},
  year={2022}
}

@inproceedings{cohen2021gradient,
  title={Gradient Descent on Neural Networks Typically Occurs at the Edge of Stability},
  author={Cohen, Jeremy M. and Kaur, Simran and Li, Yuanzhi and Kolter, J. Zico and Talwalkar, Ameet},
  booktitle={International Conference on Learning Representations},
  year={2021}
}

@article{amsel2025polar,
  title={The polar express: Optimal matrix sign methods and their application to the muon algorithm},
  author={Amsel, Noah and Persson, David and Musco, Christopher and Gower, Robert M},
  journal={arXiv preprint arXiv:2505.16932},
  year={2025}
}

@article{penedo2024fineweb,
  title={The {FineWeb} Datasets: Decanting the Web for the Finest Text Data at Scale},
  author={Penedo, Guilherme and Kydl{\'\i}{\v{c}}ek, Hynek and Lozhkov, Anton and Mitchell, Margaret and Raffel, Colin and Von Werra, Leandro and Wolf, Thomas and others},
  journal={arXiv preprint arXiv:2406.17557},
  year={2024}
}

@inproceedings{chen2025cautious,
  title         = {Cautious Weight Decay},
  author        = {Chen, Lizhang and Li, Jonathan and Liang, Kaizhao and Su, Baiyu and Xie, Cong and Pierse, Nuo Wang and Liang, Chen and Lao, Ni and Liu,
Qiang},
  booktitle     = {International Conference on Learning Representations (ICLR)},
  year          = {2026},
  eprint        = {2510.12402},
  archivePrefix = {arXiv},
  url           = {https://openreview.net/forum?id=Gwe6gbGng5}
}

@article{chen2025muon,
  title         = {{Muon} Optimizes Under Spectral Norm Constraints},
  author        = {Chen, Lizhang and Li, Jonathan and Liu, Qiang},
  journal       = {Transactions on Machine Learning Research (TMLR)},
  issn          = {2835-8856},
  year          = {2026},
  eprint        = {2506.15054},
  archivePrefix = {arXiv},
  url           = {https://openreview.net/forum?id=Blz4hjxLwU}
}

@inproceedings{liang2024memory,
  title         = {Memory-Efficient {LLM} Training with Online Subspace Descent},
  author        = {Liang, Kaizhao and Liu, Bo and Chen, Lizhang and Liu, Qiang},
  booktitle     = {Advances in Neural Information Processing Systems (NeurIPS)},
  volume        = {37},
  pages         = {64412--64432},
  year          = {2024},
  eprint        = {2408.12857},
  archivePrefix = {arXiv},
  url           = {https://proceedings.neurips.cc/paper_files/paper/2024/hash/760d09bcc06b949f5ac4b6a918739aa8-Abstract-Conference.html}
}

@inproceedings{liang2024cautious,
  title         = {Cautious Optimizers: Improving Training with One Line of Code},
  author        = {Liang, Kaizhao and Chen, Lizhang and Liu, Bo and Liu, Qiang},
  booktitle     = {International Conference on Learning Representations (ICLR)},
  year          = {2026},
  eprint        = {2411.16085},
  archivePrefix = {arXiv},
  url           = {https://openreview.net/forum?id=zBPZeRjfgu}
}

@article{dasilva2020general,
  title   = {A General System of Differential Equations to Model First-Order Adaptive Algorithms},
  author  = {Belotto da Silva, Andr{\'e} and Gazeau, Maxime},
  journal = {Journal of Machine Learning Research},
  volume  = {21},
  number  = {129},
  pages   = {1--42},
  year    = {2020}
}

@inproceedings{compagnoni2025adaptive,
  title     = {Adaptive Methods through the Lens of {SDE}s: Theoretical Insights on the Role of Noise},
  author    = {Compagnoni, Enea Monzio and Liu, Tianlin and Islamov, Rustem and Proske, Frank Norbert and Orvieto, Antonio and Lucchi, Aurelien},
  booktitle = {International Conference on Learning Representations},
  year      = {2025}
}

@article{beyn1987numerical,
  title   = {On the Numerical Approximation of Phase Portraits Near Stationary Points},
  author  = {Beyn, Wolf-J{\"u}rgen},
  journal = {SIAM Journal on Numerical Analysis},
  volume  = {24},
  number  = {5},
  pages   = {1095--1113},
  year    = {1987},
  doi     = {10.1137/0724072}
}

@book{stuart1996dynamical,
  title     = {Dynamical Systems and Numerical Analysis},
  author    = {Stuart, A. M. and Humphries, A. R.},
  series    = {Cambridge Monographs on Applied and Computational Mathematics},
  volume    = {2},
  publisher = {Cambridge University Press},
  year      = {1996}
}

@article{cohen2022adaptive,
  title   = {Adaptive Gradient Methods at the Edge of Stability},
  author  = {Cohen, Jeremy M. and Ghorbani, Behrooz and Krishnan, Shankar and Agarwal, Naman and Medapati, Sourabh and Badura, Michal and Suo, Daniel and Cardoze, David and Nado, Zachary and Dahl, George E. and Gilmer, Justin},
  journal = {arXiv preprint arXiv:2207.14484},
  year    = {2022}
}

@inproceedings{zaheer2018adaptive,
  title     = {Adaptive Methods for Nonconvex Optimization},
  author    = {Zaheer, Manzil and Reddi, Sashank J. and Sachan, Devendra and Kale, Satyen and Kumar, Sanjiv},
  booktitle = {Advances in Neural Information Processing Systems},
  pages     = {9793--9803},
  year      = {2018}
}

@inproceedings{nguyen2024memory,
  title         = {Memory-Efficient Optimization with Factorized {Hamiltonian} Descent},
  author        = {Nguyen, Son and Chen, Lizhang and Liu, Bo and Liu, Qiang},
  booktitle     = {International Conference on Artificial Intelligence and Statistics (AISTATS)},
  series        = {Proceedings of Machine Learning Research},
  volume        = {258},
  pages         = {2863--2871},
  year          = {2025},
  eprint        = {2406.09958},
  archivePrefix = {arXiv},
  url           = {https://proceedings.mlr.press/v258/nguyen25e.html}
}

@article{zhang2026adam,
  title         = {{Adam} Converges Without Any Modification On Update Rules},
  author        = {Zhang, Yushun and Li, Bingran and Chen, Congliang and Luo, Zhi-Quan and Sun, Ruoyu},
  journal       = {arXiv preprint arXiv:2603.02092},
  year          = {2026},
  eprint        = {2603.02092},
  archivePrefix = {arXiv},
  url           = {https://arxiv.org/abs/2603.02092}
}

@inproceedings{hong2024convergence,
  title         = {On Convergence of {Adam} for Stochastic Optimization under Relaxed Assumptions},
  author        = {Hong, Yusu and Lin, Junhong},
  booktitle     = {Advances in Neural Information Processing Systems (NeurIPS)},
  volume        = {37},
  year          = {2024},
  eprint        = {2402.03982},
  archivePrefix = {arXiv},
  url           = {https://arxiv.org/abs/2402.03982}
}

@article{dereich2025symmetry,
  title         = {{Adam} Symmetry Theorem: Characterization of the Convergence of the Stochastic {Adam} Optimizer},
  author        = {Dereich, Steffen and Do, Thang and Jentzen, Arnulf and von Wurstemberger, Philippe},
  journal       = {arXiv preprint arXiv:2511.06675},
  year          = {2025},
  eprint        = {2511.06675},
  archivePrefix = {arXiv},
  url           = {https://arxiv.org/abs/2511.06675}
}

@inproceedings{wang2024provable,
  title         = {Provable Adaptivity of {Adam} under Non-uniform Smoothness},
  author        = {Wang, Bohan and Zhang, Yushun and Zhang, Huishuai and Meng, Qi and Sun, Ruoyu and Ma, Zhi-Ming and Liu, Tie-Yan and Luo, Zhi-Quan and Chen, Wei},
  booktitle     = {Proceedings of the 30th ACM SIGKDD Conference on Knowledge Discovery and Data Mining (KDD)},
  year          = {2024},
  eprint        = {2208.09900},
  archivePrefix = {arXiv},
  url           = {https://arxiv.org/abs/2208.09900}
}

@article{dereich2025sharp,
  title         = {Sharp Higher Order Convergence Rates for the {Adam} Optimizer},
  author        = {Dereich, Steffen and Jentzen, Arnulf and Riekert, Adrian},
  journal       = {arXiv preprint arXiv:2504.19426},
  year          = {2025},
  eprint        = {2504.19426},
  archivePrefix = {arXiv},
  url           = {https://arxiv.org/abs/2504.19426}
}

@book{hirsch2013differential,
  title     = {Differential Equations, Dynamical Systems, and an Introduction to Chaos},
  author    = {Hirsch, Morris W. and Smale, Stephen and Devaney, Robert L.},
  edition   = {3rd},
  publisher = {Academic Press},
  year      = {2013}
}
\bibliographystyle{plainnat}

\clearpage
\appendix
\section{Deferred Proofs}

\subsection{Derivation of the AdamW ODE}

We now sketch how the AdamW ODE arises as the continuous-time limit of the discrete update, and how the discrete dynamics can be interpreted as a semi-implicit Euler scheme applied to this ODE.  

Let $\vg_t=\nabla f(\vx_{t-1})$. The AdamW updates (Algorithm~\ref{alg:adam_variants}) are
\begin{align*}
    \vm_t &= \beta_1 \vm_{t-1} + (1-\beta_1)\vg_t, \\
    \vv_t &= \beta_2 \vv_{t-1} + (1-\beta_2)\vg_t^{\odot 2}, \\
    \vx_t &= \vx_{t-1} - \eta\!\left(\frac{\vm_t}{\sqrt{\vv_t}+e} + \lambda \vx_{t-1}\right).
\end{align*}
With the common approximation $\hat\vm_t\approx \vm_t,\ \hat\vv_t\approx \vv_t$ and the substitution $\vh_t=\sqrt{\vv_t}$, we set
\[
\eta=\Delta t,\quad \beta_1=1-a\Delta t,\quad \beta_2=1-b\Delta t.
\]

\paragraph{Continuous-time limit.}  
Consider the first-moment update:
\[
\frac{\vm_t-\vm_{t-1}}{\Delta t}
= \frac{-(a\Delta t)\vm_{t-1}+(a\Delta t)\vg_t}{\Delta t}
= -a\vm_{t-1}+a\vg_t.
\]
Letting $\Delta t\to 0$, the left-hand side becomes $\dot\vm_t$, $\vm_{t-1}\to\vm_t$, and $\vg_t\to\nabla f(\vx_t)$, giving $\dot\vm_t=a(\nabla f(\vx_t)-\vm_t).$
In the same way,
\[
\frac{\vx_t-\vx_{t-1}}{\Delta t}=-\frac{\vm_t}{\vh_t+e}-\lambda\vx_{t-1}
\quad\rightarrow\quad 
\dot\vx_t=-\frac{\vm_t}{\vh_t+e}-\lambda\vx_t,
\]
and for $\vv_t=\vh_t^{\odot 2}$,
\[
\frac{\vv_t-\vv_{t-1}}{\Delta t}=-b\vv_{t-1}+b\vg_t^{\odot 2}
\quad\rightarrow\quad 
\dot\vh_t=\tfrac{b}{2}\left(\frac{\nabla f(\vx_t)^{\odot 2}}{\vh_t}-\vh_t\right).
\]

\paragraph{Semi-implicit Euler discretization.}
Starting from the AdamW ODE
\[
\dot{\vx}_t=-\frac{\vm_t}{\vh_t+e}-\lambda \vx_t,\qquad 
\dot{\vm}_t=a(\nabla f(\vx_t)-\vm_t),\qquad
\dot{\vh}_t=\tfrac{b}{2}\!\left(\tfrac{\nabla f(\vx_t)^{\odot 2}}{\vh_t}-\vh_t\right),
\]
a semi-implicit Euler step of size $\Delta t$ updates $(\vm,\vv)$ explicitly from the state at $t-1$ and then advances $\vx$ using the \emph{new} $(\vm_t,\vv_t)$. Writing $\vv=\vh^{\odot 2}$ and $\,\vg_t=\nabla f(\vx_{t-1})$:
\[
\vm_t=\vm_{t-1}+\Delta t\,a(\vg_t-\vm_{t-1})
=(1-a\Delta t)\vm_{t-1}+(a\Delta t)\vg_t,
\]
\[
\vv_t=\vv_{t-1}+\Delta t\,b(\vg_t^{\odot 2}-\vv_{t-1})
=(1-b\Delta t)\vv_{t-1}+(b\Delta t)\vg_t^{\odot 2},\quad \vh_t=\sqrt{\vv_t},
\]
\[
\vx_t
=\vx_{t-1}+\Delta t\!\left(-\frac{\vm_t}{\vh_t+e}-\lambda \vx_{t-1}\right)
=\vx_{t-1}-\eta\!\left(\frac{\vm_t}{\sqrt{\vv_t}+e}+\lambda \vx_{t-1}\right),
\]
with the identifications $\eta=\Delta t,\ \beta_1=1-a\Delta t,\ \beta_2=1-b\Delta t$. This is exactly the AdamW update without bias correction, confirming that AdamW implements a semi-implicit Euler discretization of the AdamW ODE.

\subsection{Implicit Objective of AdamW}

\begin{theorem}[Restatement of Theorem~\ref{thm:objective}]
Let $f:\sR^d\to\sR$ be continuously differentiable, and let $\lambda,e>0$. Then $\vx\in\sR^d$ satisfies the fixed-point condition~\eqref{eq:fixpoint} if and only if $\vx$ is a stationary point of the $\ell_\infty$-constrained and regularized problem
\begin{equation}
    \min_{\vx\in\sR^d}\; F(\vx)=f(\vx)+\tfrac{e}{\lambda}\,\Phi(\lambda\vx),
    \quad \text{subject to}\quad \|\vx\|_\infty\le\tfrac{1}{\lambda},
\end{equation}
where $\Phi(\vx)=\sum_{i=1}^d \phi(x_i)$ with
\[
    \phi(x)=-|x|-\log(1-|x|), \quad |x|<1.
\]
Moreover, since $\phi(x)\to+\infty$ as $|x|\to 1$, the regularizer $\Phi$ serves as a barrier function that naturally enforces the constraint $\|\vx\|_\infty<\tfrac{1}{\lambda}$.
\end{theorem}

We begin with the following lemma with the regularizer:
\begin{lemma}\label{lem:phi-basic}
Let $\phi:\R\to\R\cup\{+\infty\}$ be
\[
\phi(x)=
\begin{cases}
-|x|-\log(1-|x|), & |x|<1,\\
+\infty, & |x|\ge 1,
\end{cases}
\qquad
\Phi(\vx)=\sum_{i=1}^d\phi(x_i),
\]
and for $\lambda>0$ define the rescaled barrier $\phi_\lambda(x)\coloneqq \lambda^{-1}\phi(\lambda x)$ and $\Phi_\lambda(\vx)\coloneqq \sum_i \phi_\lambda(x_i)$. Then $\phi_\lambda$ is convex and, for $|x|<1/\lambda$,
\[
\phi_\lambda'(x)=\frac{\lambda x}{1-|\lambda x|},\qquad
\phi_\lambda''(x)=\frac{\lambda}{(1-|\lambda x|)^2}.
\]
\end{lemma}

\begin{proof}
The stated derivatives follow by direct differentiation on $(-1,1)$; convexity holds since $\phi_\lambda''\ge 0$.
\end{proof}

\begin{lemma}\label{lem:fp-eq}
For $e,\lambda>0$ and any $g,x\in\R$,
\[
\frac{g}{|g|+e}+\lambda x=0
\quad\Longleftrightarrow\quad
g+e\,\phi_\lambda'(x)=0,
\]
where the left-hand equation implies $|\lambda x|<1$, so that $\phi_\lambda'(x)$ on the right-hand side is well-defined. In $\R^d$, the equivalence holds elementwise.
\end{lemma}

\begin{proof}
Using Lemma~\ref{lem:phi-basic}, $\phi_\lambda'(x)=\frac{\lambda x}{1-|\lambda x|}$. If $\frac{g}{|g|+e}+\lambda x=0$, then $|\lambda x|<1$ and $\mathrm{sign}(x)=-\mathrm{sign}(g)$. Simple algebra yields $g+e\,\tfrac{\lambda x}{1-|\lambda x|}=0$. The converse is identical in reverse.
\end{proof}

\begin{lemma}\label{lem:grad-hess-F}
Let $F(\vx)\coloneqq f(\vx)+\frac{e}{\lambda}\Phi(\lambda \vx)=f(\vx)+e\,\Phi_\lambda(\vx)$. On $\{\,\vx:\|\lambda\vx\|_\infty<1\,\}$,
\[
\nabla F(\vx)=\nabla f(\vx)+e\,\nabla \Phi_\lambda(\vx),
\qquad
\nabla^2 F(\vx)=\nabla^2 f(\vx)+\diag\!\left(\frac{\lambda e}{(1-|\lambda \vx|)^{\odot 2}}\right).
\]
\end{lemma}

\begin{proof}
Both identities follow by differentiating coordinatewise and invoking Lemma~\ref{lem:phi-basic}.
\end{proof}

\begin{proof}[Proof of Theorem~\ref{thm:objective}]
\emph{Fixed points $\Longleftrightarrow$ stationary points.}
For AdamW fixed points (discrete or continuous limit without bias corrections), the stationary condition for $\vx$ reads
\[
\frac{\nabla f(\vx_*)}{|\nabla f(\vx_*)|+e}+\lambda \vx_*=0
\quad\text{(elementwise)}.
\]
By Lemma~\ref{lem:fp-eq} this is equivalent to
\[
\nabla f(\vx_*)+e\,\nabla \Phi_\lambda(\vx_*)=\nabla F(\vx_*)=0,
\quad
\|\lambda\vx_*\|_\infty<1.
\]
Thus fixed points lie in the open cube $\{\|\lambda\vx\|_\infty<1\}$ and are precisely the unconstrained stationary points of $F$ there.

Conversely, any stationary point of \eqref{eq:contrained} must satisfy $\|\lambda\vx_*\|_\infty<1$: since $\Phi(\lambda \vx)\to+\infty$ as $|\lambda x_i|\uparrow 1$, no optimality condition can be met on the boundary for a continuous $f$. Hence the inequality constraints are inactive, the Lagrange multipliers vanish, and the stationary conditions reduce to $\nabla F(\vx_*)=0$, which is equivalent (by Lemma~\ref{lem:fp-eq}) to the AdamW fixed-point condition. This proves the coincidence of fixed points and stationary points.
\end{proof}

\subsection{Lyapunov Function for AdamMe}\label{sec:adamme_lyapunov}

We prove Theorem~\ref{thm:adamme_lyapunov}. Throughout this subsection, all vector operations are coordinatewise, and we consider solutions initialized with
\[
    \lambda |x_i(0)|<1,\qquad v_i(0)>0,\qquad i=1,\ldots,d.
\]
On the domain $\{v_i>0\}$ the right-hand side of~\eqref{eq:adamme_ode} is locally Lipschitz, since $\nabla f$ is locally Lipschitz, $\abs{m_i}$ is Lipschitz, and $\sqrt{v_i}$ is smooth for $v_i>0$; hence solutions exist, are unique, and are $C^1$. Lemma~\ref{lem:adamme_domain_invariance} shows that this domain is preserved.
For each coordinate, define
\[
    h_i\coloneqq \sqrt{v_i},\qquad
    \hbar_i\coloneqq h_i+e,\qquad
    \psi_i(x_i,m_i)\coloneqq m_i+\lambda x_i(|m_i|+e),
\]
and let
\[
    r_i(x_i)\coloneqq -\frac{\lambda e x_i}{1-\lambda |x_i|}.
\]
Then $r_i(x_i)$ is the unique root of $\psi_i(x_i,r_i(x_i))=0$. Define the coordinate barrier induced by the same $\phi$ as in Theorem~\ref{thm:objective},
\[
    R_i(x_i)\coloneqq
    \frac{e}{\lambda}\phi(\lambda x_i),
\]
so that $R_i'(x_i)=-r_i(x_i)$, and define
\[
    J_i(x_i,m_i)\coloneqq
    \frac12\bigl(m_i|m_i|-r_i(x_i)|r_i(x_i)|\bigr)
    +e(m_i-r_i(x_i)).
\]
The coordinate energy term is
\[
I_i(x_i,m_i)
\coloneqq
\frac12(m_i^2-r_i(x_i)^2)+\lambda x_i J_i(x_i,m_i).
\]
When $b\le 4a$, consider
\[
V(\vx,\vm,\vv)
=
f(\vx)
+\sum_{i=1}^d R_i(x_i)
+\frac1a\sum_{i=1}^d\frac{I_i(x_i,m_i)}{\hbar_i}.
\]

\begin{lemma}\label{lem:adamme_lyapunov_decrease}
Along every solution of the AdamMe ODE~\eqref{eq:adamme_ode} in the above domain,
\[
    \frac{d}{dt}V(\vx(t),\vm(t),\vv(t))\le 0.
\]
\end{lemma}

\begin{proof}
Fix a coordinate $i$ and write the gradient coordinate as
\[
\begin{aligned}
    g_i &\coloneqq \partial_i f(\vx),&
    h_i &\coloneqq \sqrt{v_i},&
    \hbar_i &\coloneqq h_i+e,\\
    \psi_i &\coloneqq \psi_i(x_i,m_i),&
    r_i &\coloneqq r_i(x_i),\\
    J_i &\coloneqq J_i(x_i,m_i),&
    I_i &\coloneqq I_i(x_i,m_i).
\end{aligned}
\]
The coordinate dynamics are
\[
    \dot x_i=-\frac{\psi_i}{\hbar_i},\qquad
    \dot m_i=a(g_i-m_i),\qquad
    \dot v_i=b(g_i^2-v_i).
\]
Since $\psi_i(x_i,r_i(x_i))=0$, direct differentiation of the explicit form of $I_i$ gives
\[
    \partial_{m_i}I_i=\psi_i,\qquad
    \partial_{x_i}I_i=\lambda J_i,
\]
and
\[
    \partial_{v_i}\left(\frac{I_i}{\hbar_i}\right)
    =
    -\frac{I_i}{2h_i\hbar_i^2}.
\]
Therefore
\[
\dot V_i
=
g_i\dot x_i
+R_i'(x_i)\dot x_i
+\frac1a\frac{\partial_{x_i}I_i}{\hbar_i}\dot x_i
+\frac1a\frac{\partial_{m_i}I_i}{\hbar_i}\dot m_i
-\frac1a\frac{I_i}{2h_i\hbar_i^2}\dot v_i.
\]
Substituting the dynamics and using $R_i'(x_i)=-r_i$ gives
\[
\dot V_i
=
-\frac{g_i\psi_i}{\hbar_i}
+\frac{r_i\psi_i}{\hbar_i}
-\frac{\lambda\psi_i}{a\hbar_i^2}
J_i
+\frac{\psi_i}{\hbar_i}(g_i-m_i)
-\frac{bI_i}{2ah_i\hbar_i^2}(g_i^2-v_i).
\]
The two $g_i$ terms cancel, so
\[
\dot V_i
=
-\frac{\psi_i}{\hbar_i}
\left[
m_i-r_i
+\frac{\lambda}{a\hbar_i}J_i
\right]
-\frac{bI_i}{2ah_i\hbar_i^2}(g_i^2-v_i).
\]
Because $\lambda |x_i|<1$, the map $m\mapsto \psi_i(x_i,m)$ is strictly increasing. Hence
\[
    \operatorname{sign}(\psi_i)=\operatorname{sign}(m_i-r_i).
\]
Moreover, since $u\mapsto \frac12u|u|+eu$ is strictly increasing,
\[
    \operatorname{sign}(J_i)=\operatorname{sign}(m_i-r_i),
    \qquad
    \psi_iJ_i\ge 0.
\]
The explicit expression for $I_i$ also gives
\[
    I_i(x_i,m_i)\ge 0,
    \qquad
    \psi_i(x_i,m_i)(m_i-r_i)\ge 2I_i(x_i,m_i).
\]
Indeed, both inequalities follow by direct inspection of the three cases
$m_i,r_i\ge 0$, $m_i,r_i\le 0$, and $m_i$ and $r_i$ on opposite sides of zero.
Combining these facts,
\[
-\frac{\psi_i}{\hbar_i}
\left[
m_i-r_i
+\frac{\lambda}{a\hbar_i}J_i
\right]
\le
-\frac{2I_i}{\hbar_i}.
\]
Therefore
\[
    \dot V_i
    \le
    -\frac{2I_i}{\hbar_i}
    -\frac{bI_i}{2ah_i\hbar_i^2}(g_i^2-v_i).
\]
If $g_i^2\ge v_i$, the second term is nonpositive. If $g_i^2<v_i$, then
\[
-\frac{bI_i}{2ah_i\hbar_i^2}(g_i^2-v_i)
=
\frac{bI_i}{2ah_i\hbar_i^2}(v_i-g_i^2)
\le
\frac{bI_i h_i}{2a\hbar_i^2},
\]
and hence
\[
\dot V_i
\le
-\frac{I_i}{\hbar_i}
\left[
2-\frac{b h_i}{2a\hbar_i}
\right].
\]
Since $0<h_i/\hbar_i<1$ and $b\le 4a$, the bracket is nonnegative. Thus $\dot V_i\le 0$ in both cases. Summing over coordinates proves the claim.
\end{proof}

\begin{lemma}\label{lem:adamme_domain_invariance}
The domain $\{\vx:\lambda |x_i|<1,\ i=1,\ldots,d\}$ is forward invariant. Moreover, if $v_i(0)>0$, then $v_i(t)>0$ for all $t\ge 0$.
\end{lemma}

\begin{proof}
At the boundary $x_i=1/\lambda$,
\[
    \psi_i=m_i+|m_i|+e>0,
\]
so $\dot x_i=-\psi_i/(\sqrt{v_i}+e)<0$. At the boundary $x_i=-1/\lambda$,
\[
    \psi_i=m_i-|m_i|-e<0,
\]
so $\dot x_i=-\psi_i/(\sqrt{v_i}+e)>0$. Thus the vector field points inward at the boundary of the box.

For the second claim, the $v_i$ equation gives
\[
    v_i(t)
    =
    e^{-bt}v_i(0)
    +
    b\int_0^t e^{-b(t-s)}
    \bigl(\partial_i f(\vx(s))\bigr)^2\,ds
    >0.
\]
\end{proof}

\begin{proof}[Proof of Theorem~\ref{thm:adamme_lyapunov}]
Lemmas~\ref{lem:adamme_lyapunov_decrease} and~\ref{lem:adamme_domain_invariance} show that $V$ is a Lyapunov function on the stated forward-invariant domain. For any bounded trajectory, the remaining argument follows the standard LaSalle routine, as in the final step of Theorem~B.3 of~\citet{chen2023lion}. Let the limiting set denote the set of all subsequential limits along times $t_k\to\infty$. LaSalle's invariance principle implies that this limiting set is contained in the largest invariant subset of $\dot V=0$, with our vector field and Lyapunov function replacing the Lion-$\mathcal K$ dynamics and Lyapunov function there.

From the proof of Lemma~\ref{lem:adamme_lyapunov_decrease}, equality requires $I_i(x_i,m_i)=0$ for every coordinate, hence $m_i=r_i(x_i)$ and $\psi_i(x_i,m_i)=0$. Therefore $\dot{\vx}=0$. On any invariant subset of this set, we must also have $\dot{\vm}=0$, so $\vm=\nabla f(\vx)$. Finally, $\dot{\vv}=0$ gives $\vv=\nabla f(\vx)^{\odot 2}$. These are exactly the equilibrium conditions for~\eqref{eq:adamme_ode}. Hence every bounded trajectory approaches the equilibrium set. If the limiting equilibria are isolated, the trajectory converges to a single equilibrium.

\end{proof}

\subsection{Discrete-Time Convergence of AdamMe}\label{sec:adamme_discrete}

We adapt the energy-increment argument of Theorem~B.8 and the convex-combination argument of Theorem~B.10 of~\citet{chen2023lion}. Those results concern Lion-$\mathcal K$ with a fixed convex potential; the additional step for AdamMe is to control its changing second-moment denominator. The functions $F$, $I_i$, $\psi_i$, and $r_i$ below are those defined above.

\begin{proposition}\label{prop:adamme_discrete}
Assume $f$ is $C^1$ with $L$-Lipschitz gradient in the Euclidean norm on a neighborhood of $\mathcal C=\{\|\lambda\vx\|_\infty\le1\}$, $\lambda,e>0$, and $\beta_1,\beta_2\in(0,1)$ satisfy
\begin{equation}\label{eq:adamme_discrete_momentum}
    s\coloneqq\sqrt{\beta_2}>\frac{\beta_1}{2-\beta_1}.
\end{equation}
Given an initialization with $\|\lambda\vx_0\|_\infty<1$ and $\vv_0\ge0$, write $\alpha=1-\beta_1$ and define
\[
\begin{aligned}
M&=\max\left\{\|\vm_0\|_\infty,\ \|\nabla f(\vzero)\|_\infty+\frac{L\sqrt d}{\lambda}\right\},\\
\rho&=\min\left\{1-\|\lambda\vx_0\|_\infty,\ \frac{e}{M+e}\right\},
\qquad \kappa=2-\frac{\beta_1(1-s)}{\alpha s}>0,\\
L_0&=L+\frac{\lambda e}{\rho^2},\qquad
L_1=\frac{\beta_1\lambda^2e}{\alpha s\rho^3}.
\end{aligned}
\]
For every constant step size
\begin{equation}\label{eq:adamme_discrete_stepsize}
0<\eta\le\eta_{\max}\coloneqq
\min\left\{\frac{e}{\lambda(M+e)},\quad
\frac{\kappa e}{2\left(L_0+\sqrt{L_0^2+\kappa eL_1}\right)}\right\},
\end{equation}
deterministic AdamMe without bias correction remains in $\{\|\lambda\vx\|_\infty\le1-\rho\}$ and satisfies
\[
    \frac1T\sum_{k=0}^{T-1}\|\nabla F(\vx_k)\|^2=O(T^{-1}),
    \qquad \nabla F(\vx_k)\longrightarrow0.
\]
If the stationary points in this compact set are isolated, the full state converges to an equilibrium. The same conclusions hold with the bias corrections and zero moment initialization in Algorithm~\ref{alg:adam_variants}.
\end{proposition}

\begin{proof}
Write $d_{k,i}=\sqrt{v_{k,i}}+e$ and $\Delta m_{k,i}=m_{k+1,i}-m_{k,i}$. The updates are
\[
\begin{aligned}
\vm_{k+1}&=\beta_1\vm_k+\alpha\nabla f(\vx_k),\qquad
\vv_{k+1}=\beta_2\vv_k+(1-\beta_2)\nabla f(\vx_k)^{\odot2},\\
x_{k+1,i}&=x_{k,i}-\eta\frac{\psi_i(x_{k,i},m_{k+1,i})}{d_{k+1,i}}.
\end{aligned}
\]
Smoothness gives $\|\nabla f(\vx)\|_\infty\le\|\nabla f(\vzero)\|_\infty+L\|\vx\|_2\le M$ on $\mathcal C$.
Set $q=1-\rho=\max\{\|\lambda\vx_0\|_\infty,M/(M+e)\}<1$.
The first bound in~\eqref{eq:adamme_discrete_stepsize} ensures $\eta\lambda(M+e)\le e$, so the identity
\[
\lambda x_{k+1,i}=(1-\theta_{k,i})\lambda x_{k,i}
-\theta_{k,i}\frac{m_{k+1,i}}{|m_{k+1,i}|+e},
\qquad
\theta_{k,i}=\eta\lambda\frac{|m_{k+1,i}|+e}{d_{k+1,i}}\in[0,1]
\]
proves inductively that $\|\lambda\vx_k\|_\infty\le q$, as in the proof of Theorem~B.10 of~\citet{chen2023lion}. Both moments are then bounded, and $e\le d_{k,i}\le D$ for some $D<\infty$.

Define the discrete energy by
\[
W_k=F(\vx_k)+c_\eta\sum_i\frac{I_i(x_{k,i},m_{k,i})}{d_{k,i}},
\qquad c_\eta=\frac{\eta\beta_1}{\alpha s}.
\]
The second-moment update gives $d_{k+1,i}\ge s d_{k,i}$. Moreover, $I_i(x,\cdot)$ is $\rho$-strongly convex with derivative $\psi_i(x,\cdot)$, whose Lipschitz constant is at most $2$. Thus, abbreviating $I_i^+=I_i(x_{k,i},m_{k+1,i})$ and $\psi_i^+=\psi_i(x_{k,i},m_{k+1,i})$, we obtain the required denominator estimate:
\[
\frac{I_i^+}{d_{k+1,i}}-\frac{I_i(x_{k,i},m_{k,i})}{d_{k,i}}
\le
\frac{(1-s)I_i^++s\psi_i^+\Delta m_{k,i}
-\tfrac{s\rho}{2}(\Delta m_{k,i})^2}{d_{k+1,i}}.
\]
For the position update, hold the updated moments $(\vm_{k+1},\vv_{k+1})$ fixed and regard the energy as a function of $\vx$ alone:
\[
P_k(\vx)\coloneqq F(\vx)+c_\eta\sum_i\frac{I_i(x_i,m_{k+1,i})}{d_{k+1,i}}.
\]
The full energy increment splits into a position increment and a moment increment,
\[
W_{k+1}-W_k
=P_k(\vx_{k+1})-P_k(\vx_k)
+c_\eta\sum_i\left(\frac{I_i^+}{d_{k+1,i}}-\frac{I_i(x_{k,i},m_{k,i})}{d_{k,i}}\right).
\]
The preceding denominator estimate controls the second term. To bound the first, note that the barrier in $F$ contributes at most $\lambda e/\rho^2$ to its gradient Lipschitz constant. Since
$\partial_{x_i}I_i=\lambda J_i$, $\partial_{x_i}^2I_i=\lambda^2e^2/(1-\lambda|x_i|)^3$, and $d_{k+1,i}\ge e$, the function $P_k$ is $L_\eta$-smooth on the invariant cube, where
\[
L_\eta=L+\frac{\lambda e}{\rho^2}+\frac{c_\eta\lambda^2e}{\rho^3}=L_0+L_1\eta.
\]
Write $g_{k,i}=\nabla_i f(\vx_k)$, $r_{k,i}=r_i(x_{k,i})$, and $J_i^+=J_i(x_{k,i},m_{k+1,i})$. Then
$\partial_iP_k(\vx_k)=g_{k,i}-r_{k,i}+c_\eta\lambda J_i^+/d_{k+1,i}$.
Substituting $\Delta x_{k,i}=x_{k+1,i}-x_{k,i}=-\eta\psi_i^+/d_{k+1,i}$ into the smoothness inequality gives
\[
\begin{aligned}
P_k(\vx_{k+1})-P_k(\vx_k)
&\le\sum_i\partial_iP_k(\vx_k)\Delta x_{k,i}
+\frac{L_\eta}{2}\|\Delta\vx_k\|^2\\
&=-\eta\sum_i\frac{(g_{k,i}-r_{k,i})\psi_i^+}{d_{k+1,i}}
-c_\eta\eta\lambda\sum_i\frac{\psi_i^+J_i^+}{d_{k+1,i}^2}
+\frac{L_\eta\eta^2}{2}\sum_i\frac{(\psi_i^+)^2}{d_{k+1,i}^2}.
\end{aligned}
\]
Adding the moment-increment bound makes the cancellation explicit: because
$g_{k,i}=m_{k+1,i}+\beta_1\Delta m_{k,i}/\alpha$ and $c_\eta s=\eta\beta_1/\alpha$,
\[
-\eta\frac{(g_{k,i}-r_{k,i})\psi_i^+}{d_{k+1,i}}
+c_\eta s\frac{\psi_i^+\Delta m_{k,i}}{d_{k+1,i}}
=-\eta\frac{(m_{k+1,i}-r_{k,i})\psi_i^+}{d_{k+1,i}}.
\]
Finally, $\psi_i^+J_i^+\ge0$ makes the second term in the position bound nonpositive, while
$\psi_i^+(m_{k+1,i}-r_{k,i})\ge2I_i^+$ bounds the remaining linear term.
The quadratic remainder satisfies
\[
\frac{L_\eta\eta^2}{2}\sum_i\frac{(\psi_i^+)^2}{d_{k+1,i}^2}
\le\frac{2L_\eta\eta^2}{e}\sum_i\frac{I_i^+}{d_{k+1,i}},
\]
using $(\psi_i^+)^2\le4I_i^+$ and $d_{k+1,i}\ge e$.
Combining these bounds with $2\eta-c_\eta(1-s)=\eta\kappa$ yields
\begin{equation}\label{eq:adamme_discrete_descent}
W_{k+1}-W_k
\le-\eta\left(\kappa-\frac{2\eta L_\eta}{e}\right)S_k
-\frac{\eta\beta_1\rho}{2\alpha}Q_k,
\end{equation}
where $S_k=\sum_i I_i^+/d_{k+1,i}$ and $Q_k=\sum_i(\Delta m_{k,i})^2/d_{k+1,i}$.
The second bound in~\eqref{eq:adamme_discrete_stepsize} is the positive root bound for $L_1\eta^2+L_0\eta\le\kappa e/4$.
Since $L_\eta=L_0+L_1\eta$, this ensures $2\eta L_\eta/e\le\kappa/2$, leaving a strictly positive coefficient for both $S_k$ and $Q_k$ in the descent estimate.
Let $F_{\min}=\min_{\|\lambda\vx\|_\infty\le q}F(\vx)$, which is finite by continuity on the compact invariant cube. Because $I_i\ge0$, we have $W_k\ge F_{\min}$. Summing~\eqref{eq:adamme_discrete_descent} from $k=0$ to $T-1$ therefore gives
\[
\frac{\eta\kappa}{2}\sum_{k=0}^{T-1}S_k
+\frac{\eta\beta_1\rho}{2\alpha}\sum_{k=0}^{T-1}Q_k
\le W_0-W_T\le W_0-F_{\min}.
\]
The right-hand side is independent of $T$, so both nonnegative sequences are summable; in particular, $S_k\to0$ and $Q_k\to0$.

To translate these energy residuals into stationarity, use the strong convexity of $I_i(x_{k,i},\cdot)$ around its minimizer $r_i(x_{k,i})$:
\[
I_i^+\ge\frac{\rho}{2}|m_{k+1,i}-r_i(x_{k,i})|^2,
\qquad
\nabla_i F(\vx_k)=m_{k+1,i}-r_i(x_{k,i})+\frac{\beta_1}{\alpha}\Delta m_{k,i}.
\]
Applying $\|u+w\|^2\le2\|u\|^2+2\|w\|^2$ and the upper bound $d_{k+1,i}\le D$ yields
\[
\|\nabla F(\vx_k)\|^2
\le\frac{4D}{\rho}S_k+\frac{2\beta_1^2D}{\alpha^2}Q_k.
\]
Combining this inequality with the preceding telescoping bound gives, for every $T$,
\[
\sum_{k=0}^{T-1}\|\nabla F(\vx_k)\|^2
\le\frac{D(W_0-F_{\min})}{\eta\rho}
\max\left\{\frac{8}{\kappa},\frac{4\beta_1}{\alpha}\right\}.
\]
Thus the squared gradients are summable and $\nabla F(\vx_k)\to0$. Dividing the same bound by $T$ proves the claimed $O(T^{-1})$ average stationarity rate.

For convergence of the iterates, the position update also gives
\[
\|\vx_{k+1}-\vx_k\|^2
=\eta^2\sum_i\frac{(\psi_i^+)^2}{d_{k+1,i}^2}
\le\frac{4\eta^2}{e}S_k\longrightarrow0.
\]
Since the iterates remain in a compact set and $\nabla F(\vx_k)\to0$, these vanishing increments imply convergence to a stationary point $\vx_*$ whenever the stationary points are isolated.

Since $S_k\to0$, strong convexity yields $\vm_{k+1}-\vr(\vx_k)\to0$. Thus $\vm_k\to\vr(\vx_*)=\vg_*$, where $\vg_*=\nabla f(\vx_*)$ by stationarity. The second-moment error obeys
\[
\vv_{k+1}-\vg_*^{\odot2}
=\beta_2(\vv_k-\vg_*^{\odot2})
+(1-\beta_2)\bigl(\nabla f(\vx_k)^{\odot2}-\vg_*^{\odot2}\bigr).
\]
Because $0<\beta_2<1$ and the forcing term vanishes, this recursion gives $\vv_k\to\vg_*^{\odot2}$. Hence the full state converges to an equilibrium.

Bias-corrected moments are bounded convex averages of the past gradients and their squared values, preserving domain invariance. Their position update differs from the uncorrected update by $O(\eta(\beta_1^{k+1}+\beta_2^{k+1}))$ on this compact set. This adds a summable remainder to~\eqref{eq:adamme_discrete_descent}, preserving the convergence conclusions.
\end{proof}

The condition in~\eqref{eq:adamme_discrete_momentum} is the discrete counterpart of $b<4a$: substituting $\beta_1=1-a\eta$ and $\beta_2=1-b\eta$ gives $\kappa\to2-b/(2a)$ as $\eta\to0$.

\paragraph{Stochastic gradients.}
The conditional-energy argument parallels Theorem~B.14 of~\citet{chen2023lion} (version~7). Assume additionally that the sampled gradients satisfy $\|\vg_k\|_\infty\le G$ on the cube, are conditionally unbiased, and have conditional variance at most $\sigma^2/B_k$, where $B_k$ is the batch size. Set $M=\max\{\|\vm_0\|_\infty,G\}$ and compute $\rho,L_0,L_1,\eta_{\max}$ as in Proposition~\ref{prop:adamme_discrete}. Writing $\vg_k=\nabla f(\vx_k)+\boldsymbol{\xi}_k$, the only additional term in~\eqref{eq:adamme_discrete_descent} is $\eta\sum_i\psi_i^+\xi_{k,i}/d_{k+1,i}$.
Young's inequality and $(\psi_i^+)^2\le4I_i^+$ bound this by $\eta\kappa S_k/4+4\eta\|\boldsymbol{\xi}_k\|^2/(\kappa e)$. Therefore, for every $0<\eta\le\eta_{\max}$,
\begin{gather*}
\mathbb E_k W_{k+1}\le W_k-\frac{\eta\kappa}{4}\mathbb E_k S_k
-\frac{\eta\beta_1\rho}{2\alpha}\mathbb E_k Q_k
+\frac{4\eta\sigma^2}{\kappa e B_k},\\
\frac1T\sum_{k<T}\mathbb E\|\nabla F(\vx_k)\|^2
\le\frac{C_0}{\eta T}+\frac{C_1\sigma^2}{T}\sum_{k<T}B_k^{-1},
\end{gather*}
where $\mathbb E_k$ conditions on the history before sampling $\vg_k$ and $C_0,C_1$ are independent of $T$ and the batch-size schedule. The second bound follows by telescoping and using
$\nabla F(\vx_k)=\vm_{k+1}-\vr(\vx_k)+\beta_1(\vm_{k+1}-\vm_k)/\alpha-\boldsymbol{\xi}_k$.
Thus fixed batches yield an $O(T^{-1}+\sigma^2/B)$ stationarity bound. If $\sum_k B_k^{-1}<\infty$, telescoping the first bound gives $\sum_k(S_k+Q_k+\|\boldsymbol{\xi}_k\|^2)<\infty$ almost surely, hence $\nabla F(\vx_k)\to0$ almost surely, and convergence to one equilibrium when stationary points are isolated. Bias correction again contributes only a summable remainder.

\subsection{Stability Condition of AdamW ODE}
Writing the AdamW ODE in a vectorized form, we have:
\begin{equation}
\frac{d}{dt} \vz = \mT(\vz), \quad
\vz \coloneqq \begin{bmatrix}\vx\\\vm\\\vh\end{bmatrix},\quad
\mT(\vz) \coloneqq
\begin{bmatrix}
-\vm/(\vh+e) -\lambda\vx \\
a(\nabla f(\vx) - \vm) \\
\frac{b}{2}(\nabla f(\vx)^{\odot 2}/\vh - \vh)
\end{bmatrix}.
\end{equation}
By differentiating the system $\mJ(\vz)\coloneqq \partial\mT(\vz)/\partial \vz$, we obtain:
\[
\mJ(\vz) =
\begin{bmatrix}
-\lambda \mI & -\diag(1/(\vh+e)) & \diag(\vm/(\vh+e)^2) \\
a\nabla^2 f(\vx) & -a \mI & 0 \\
b\diag(\nabla f(\vx)/\vh)\nabla^2 f(\vx) & 0 & -\tfrac{b}{2}\bigl(\diag(\nabla f(\vx)^{\odot 2}/\vh^2)+\mI\bigr)
\end{bmatrix}.
\]
Noting that at equilibrium point $\vz_* = [\vx_*, \vm_*, \vh_*]$, we have $\nabla f(\vx_*)^{\odot 2} = \vh_*^{\odot 2}$, this yields:
\[
    \mJ(\vz_*)=
    \begin{bmatrix}
    -\lambda\mI & -\diag(1/(\vh_*+e)) & \diag(\vm_*/(\vh_*+e)^2)\\
    a\nabla^2 f(\vx_*) & -a\mI & 0\\
    b\diag(\nabla f(\vx_*)/\vh_*)\nabla^2 f(\vx_*) & 0 & -b\mI
    \end{bmatrix}.
\]
This following lemma is useful for simplifying the characteristic polynomial:
\begin{lemma}\label{lem:hessianF}
If $[\vx_*,\vm_*,\vh_*]$ is a fixed point of AdamW in Eq.~\eqref{eq:fixpoint}, then
\[
\nabla^2 F(\vx_*) 
= \nabla^2 f(\vx_*) + \diag{\tfrac{\lambda e}{(1-\lvert \lambda \vx_* \rvert)^2}}
= \nabla^2 f(\vx_*) + \diag{\tfrac{\lambda}{e}\,(\vh_*+e)^2}.
\]
\end{lemma}

\begin{proof}
At equilibrium, we have $\frac{\vm_*}{\vh_*+e} + \lambda \vx_* = 0$ and $\vh_* = \lvert \nabla f(\vx_*)\rvert = \lvert \vm_* \rvert$. Hence
\[
\lvert \lambda \vx_* \rvert
= \frac{\lvert \vm_* \rvert}{\vh_*+e}
= \frac{\vh_*}{\vh_*+e}.
\]
Therefore,
\[
\frac{\lambda e}{(1-\lvert \lambda \vx_* \rvert)^2}
= \frac{\lambda e}{(e/(\vh_*+e))^2}
= \frac{\lambda}{e}\,(\vh_*+e)^2.
\]
The claim then follows from Lemma~\ref{lem:grad-hess-F}.
\end{proof}

\begin{lemma}[Restatement of Lemma~\ref{lem:rootcubiceq}]
All roots of the cubic equation $s^3 + \alpha_2 s^2 + \alpha_1 s + \alpha_0 =0$ have negative real part (i.e., the system is Hurwitz stable) if and only if
$$
\alpha_2,\ \alpha_1,\ \alpha_0 > 0 \quad \text{and} \quad \alpha_2 \alpha_1 > \alpha_0.
$$
\end{lemma}

\begin{proof}[Proof of Lemma~\ref{lem:rootcubiceq}]
($\Rightarrow$)  
Suppose all roots satisfy $\Re(s_i)<0$. Writing the polynomial as 
\[
(s-s_1)(s-s_2)(s-s_3) = s^3 + \alpha_2 s^2 + \alpha_1 s + \alpha_0,
\]
Vieta's relations give
\[
\alpha_2 = -(s_1+s_2+s_3), \quad 
\alpha_1 = s_1s_2+s_1s_3+s_2s_3, \quad 
\alpha_0 = -s_1s_2s_3.
\]
The polynomial factors into terms $s-r$ with $-r>0$ for real roots $r$, and $(s-p)(s-\bar p)=s^2-2\Re(p)\,s+|p|^2$ with $-2\Re(p)>0$ for complex-conjugate pairs; all of these have positive coefficients, so $\alpha_2,\alpha_1,\alpha_0>0$. Moreover, direct expansion gives
\[
\alpha_2 \alpha_1 - \alpha_0 
= -(s_1+s_2)(s_1+s_3)(s_2+s_3).
\]
If all roots are real, each factor $s_i+s_j$ is negative, so the right-hand side is positive. If $s_1=p$, $s_2=\bar p$, and $s_3=r$ is real, then $s_1+s_2=2\Re(p)<0$ and $(s_1+s_3)(s_2+s_3)=|p+r|^2>0$, where $p+r\neq0$ because $\Re(p+r)<0$; so the right-hand side is again positive. Hence $\alpha_2\alpha_1>\alpha_0$.

($\Leftarrow$)  
Conversely, assume $\alpha_2,\alpha_1,\alpha_0>0$ and $\alpha_2\alpha_1>\alpha_0$. If $s\in\mathbb{R}$ satisfies $s\geq 0$, then 
\[
s^3 + \alpha_2 s^2 + \alpha_1 s + \alpha_0 \geq \alpha_0 > 0,
\]
so $s$ is not a root. Hence all real roots satisfy $s<0$.  
Now let $s=u+iv$ with $v\neq 0$. Substituting and separating real and imaginary parts yields
\[
3u^2 + 2\alpha_2 u + \alpha_1 = v^2, 
\qquad 
u^3 + \alpha_2 u^2 + \alpha_1 u + \alpha_0 = (3u+\alpha_2)v^2.
\]
Eliminating $v^2$ gives
\[
8u^3 + 8\alpha_2 u^2 + 2(\alpha_2^2+\alpha_1)u + (\alpha_2\alpha_1-\alpha_0) = 0.
\]
If $u\geq 0$, the left-hand side is at least $\alpha_2\alpha_1-\alpha_0>0$, a contradiction. Thus $u<0$, i.e., all complex roots lie in the open left half-plane.

Therefore, the given conditions are necessary and sufficient for Hurwitz stability.
\end{proof}

\begin{proposition}[Restatement of Proposition~\ref{prop:stable_iff}]
Assume $a,b,e>0$ and $\lambda\ge0$. Let $[x_*,m_*,h_*]$ be an equilibrium of the AdamW ODE. Then $[x_*,m_*,h_*]$ is Hurwitz stable if and only if $F''(x_*)>0$ and 
\begin{equation}
[a (a + \lambda) - b (b + \lambda) \abs{\lambda x_*}] f'' (x_*)  > - \frac{e C}{(1-\abs{\lambda x_*})},
\end{equation}
where $C = (a+b+\lambda)(ab+a\lambda+b\lambda) - ab\lambda > 0$.
\end{proposition}

\begin{proof}[Proof of Proposition~\ref{prop:stable_iff}]
Assume $a,b,e>0$ and $\lambda \geq 0$. Recall from the Routh--Hurwitz criterion (Lemma~\ref{lem:rootcubiceq}) that all roots of a cubic $s^3 + \alpha_2 s^2 + \alpha_1 s + \alpha_0$ have negative real part if and only if
\[
\alpha_2,\ \alpha_1,\ \alpha_0 > 0
\quad \text{and} \quad 
\alpha_2 \alpha_1 > \alpha_0.
\]

For the characteristic polynomial of $\mJ(\vz_*)$,
\[
\chi_\mJ(s) 
= s^3 + (a+b+\lambda) s^2 
+ \left[ab + a\lambda + \frac{(a-b)H_*}{\hbar_*} + \frac{be \Hat{H}_*}{\hbar_*^2}\right] s 
+ \frac{abe \Hat{H}_*}{\hbar_*^2},
\]
we identify
\[
\alpha_2 = a+b+\lambda,\quad 
\alpha_1 = ab + a\lambda + \frac{(a-b)H_*}{\hbar_*} + \frac{be \Hat{H}_*}{\hbar_*^2},\quad 
\alpha_0 = \frac{abe\Hat{H}_*}{\hbar_*^2}.
\]

Since $a,b,e>0$ and $\lambda \ge 0$, clearly $\alpha_2>0$. Moreover, $\alpha_0>0$ if and only if $\Hat{H}_* = F''(x_*)>0$, which shows that any Hurwitz stable equilibrium must be a strict local minimum of $F$.

It remains to impose $\alpha_2 \alpha_1 > \alpha_0$; together with $\alpha_0>0$ and $\alpha_2>0$, this gives $\alpha_1>\alpha_0/\alpha_2>0$, so the condition $\alpha_1>0$ holds automatically. Substituting the definitions and rearranging, one obtains
\[
\bigl[(a+b+\lambda)(a\hbar_* - b h_*) - abe\bigr] H_* 
> -\bigl[ (a+b+\lambda)(ab+a\lambda+b\lambda) - ab\lambda \bigr]\hbar_*^2.
\]
Dividing by $\hbar_*>0$ and using $h_*/\hbar_* = \lvert \lambda x_* \rvert$ and $\hbar_* = e/(1-\lvert \lambda x_* \rvert)$, this inequality is equivalent to
\[
\bigl[a(a+\lambda) - b(b+\lambda)\lvert \lambda x_* \rvert\bigr] f''(x_*) 
> -\frac{e C}{1-\lvert \lambda x_* \rvert},
\]
where 
\[
C \coloneqq (a+b+\lambda)(ab+a\lambda+b\lambda) - ab\lambda > 0.
\]

Thus the equilibrium $(x_*,m_*,h_*)$ is Hurwitz stable if and only if $F''(x_*)>0$ and the above inequality holds.
\end{proof}

\begin{corollary}\label{cor:vanish}
Assume $a, b, e>0$, and $\lambda \geq 0$. Let $[x_*, m_*, h_*]$ be an equilibrium point of AdamW ODE. If $a = b$, or $\lambda = 0$, then $[x_*, m_*, h_*]$ is Hurwitz stable iff $F''(x_*) > 0$.   
\end{corollary}

\begin{proof}
By Lemma~\ref{lem:rootcubiceq}, Hurwitz stability of an equilibrium $\vz_*=[x_*,m_*,h_*]$ requires $F''(x_*)>0$ and
\[
(a+b+\lambda)\left[ab+a\lambda+\frac{(a-b)H_*}{\hbar_*}+\frac{be F''(x_*)}{\hbar_*^2}\right] > \frac{abe F''(x_*)}{\hbar_*^2}.
\]
This can be rearranged as
\[
(a+b+\lambda)\left[ab+a\lambda+\frac{(a-b)H_*}{\hbar_*}\right] + (b+\lambda)\frac{be F''(x_*)}{\hbar_*^2} > 0.
\]

If $a=b$, the term involving $(a-b)$ vanishes, and the inequality simplifies to
\[
(a+b+\lambda)(ab+a\lambda) + (b+\lambda)\frac{be F''(x_*)}{\hbar_*^2} > 0.
\]
Since $a=b>0$, $\lambda \ge 0$, and $e>0$, this condition is automatically satisfied whenever $F''(x_*)>0$.

If $\lambda=0$, the dynamics reduce to the unconstrained Adam system. In this case $\hbar_*=e$, $h_*=0$, and $F''(x_*)=H_*$. The inequality then becomes
\[
(a+b)\left[ab+\frac{(a-b)H_*}{e}\right] + \frac{b^2 H_*}{e} > 0,
\]
which simplifies further to
\[
(a+b)ab + \frac{a^2}{e}H_* > 0.
\]
Because $a,b,e>0$, this condition is implied by $H_*=F''(x_*)>0$.

In both cases, stability is therefore equivalent to $F''(x_*)>0$, completing the proof.
\end{proof}

\subsection{The Discrete AdamW System}\label{sec:discrete}

We consider exact gradients and the autonomous AdamW update without bias correction, as in~\secref{sec:nonconverge}. The calculation in $h=\sqrt v$ coordinates applies at active fixed points, where $h_*=|f'(x_*)|>0$. We first derive the Jacobian and then express its Schur conditions as curvature bounds that recover the continuous-time analysis.

\begin{definition}[Lyapunov stability for discrete systems]
Consider the autonomous system $\vz_{t+1} = \mT(\vz_t)$, where $\mT:\gD \to \sR^{3d}$ with domain $\gD \subset \sR^{3d}$.  
Let $\vz_* \in \gD$ be a fixed point, i.e. $\mT(\vz_*) = \vz_*$. Then:
\begin{itemize}
  \item $\vz_*$ is \textbf{Lyapunov stable} if for every $\epsilon > 0$ there exists $\delta > 0$ such that $\norm{\vz_0 - \vz_*} < \delta$ implies $\norm{\vz_t - \vz_*} < \epsilon$ for all integers $t \ge 0$.
  \item $\vz_*$ is \textbf{asymptotically stable} if it is Lyapunov stable and there exists $\delta' > 0$ such that $\norm{\vz_0 - \vz_*} < \delta'$ implies $\lim_{t \to \infty} \vz_t = \vz_*$.
  \item $\vz_*$ is \textbf{unstable} if it is not Lyapunov stable; i.e., there exists $\epsilon' > 0$ such that for every $\delta > 0$ one can find $\norm{\vz_0 - \vz_*} < \delta$ but $\norm{\vz_t - \vz_*} \ge \epsilon'$ for some integer $t > 0$.
\end{itemize}
\end{definition}

Lyapunov’s stability characterizes robustness of discrete trajectories, while the following Schur stability criterion describes how the linearized neighborhood near a fixed point behaves.

\begin{definition}[Schur stability]
Let $\vz_*$ be a fixed point of $\vz_{t+1} = \mT(\vz_t)$ and let $\mJ(\vz_*) = \bigl[\partial T_i / \partial z_j\bigr]_{\vz_*}$ denote the Jacobian at $\vz_*$.  
Then $\vz_*$ is \textbf{Schur stable} if every eigenvalue $s \in \sC$ of $\mJ(\vz_*)$ satisfies $|s| < 1$.
\end{definition}

\emph{Lyapunov’s indirect method (discrete form)} states: If $\mJ(\vz_*)$ is Schur stable, then the fixed point $\vz_*$ is \emph{asymptotically stable}. If $\mJ(\vz_*)$ has an eigenvalue with $|s| > 1$, the fixed point is \emph{unstable}. 

For the Schur-stability calculation, we work with the first-order local representation of the $h=\sqrt v$ dynamics around a fixed point:
\begin{equation}
\vz_{t + 1} = \mT(\vz_t), \quad
\vz \coloneqq \begin{bmatrix}\vx\\\vm\\\vh\end{bmatrix},\quad
\mT(\vz) \coloneqq
\begin{bmatrix}
(1 - \lambda\eta)\vx - \eta \frac{\beta_1 \vm + (1 - \beta_1) \nabla f(\vx)}{\beta_2' \vh + (1 - \beta_2') \nabla f(\vx)^{\odot 2} / \vh + e}  \\
\beta_1 \vm + (1 - \beta_1) \nabla f(\vx) \\
\beta_2' \vh + (1 - \beta_2') \nabla f(\vx)^{\odot 2}/\vh
\end{bmatrix}.
\end{equation}

where $\beta_2' = \frac{1 + \beta_2}{2}$. This representation has the same fixed point and Jacobian as the exact AdamW $v$-update. Note that the term corresponding to $\vx_{t+1}$ involves more terms than the continuous counterpart. This is because AdamW update corresponds to the semi-implicit Euler discretization of the AdamW ODE and we need to write $\vx_{t+1}$ in terms of $[\vx_t, \vm_t, \vh_t]$ only for analysis. Nevertheless, we will use $\vm_{t+1}$ and $\vh_{t+1}$ to simplify expressions in the following. Differentiating this system $\mJ(\vz) \coloneqq \partial \mT(\vz) / \partial \vz$, we obtain:

\[
    \mJ(\vz_t) =
    \begin{bmatrix}
    \frac{\partial \vx_{t+1}}{\partial \vx_t} & -\eta\beta_1 \diag(\frac{1}{\vh_{t+1}+e}) & \frac{\partial \vx_{t+1}}{\partial \vh_t} \\
    (1 - \beta_1) \nabla^2 f(\vx_t) & \beta_1 \mI & 0 \\
    (1 - \beta_2) \diag(\frac{\nabla f(\vx_t)}{\vh_t}) \nabla^2 f(\vx_t) & 0 & (1 + \beta_2) \mI - \diag(\frac{\vh_{t+1}}{\vh_t})
    \end{bmatrix}.
\]
where
\begin{align*}
    \frac{\partial \vx_{t+1}}{\partial \vx_t} &= (1 - \lambda\eta) \mI - \eta (1 - \beta_1) \diag\left(\frac{1}{\vh_{t+1}+e}\right)\nabla^2 f(\vx_t) \\ 
    &\quad + \eta (1 - \beta_2) \diag\left( \frac{\vm_{t+1} \nabla f(\vx_t)}{\vh_t (\vh_{t+1}+e)^2} \right) \nabla^2 f(\vx_t) \\
    \frac{\partial \vx_{t+1}}{\partial \vh_t} &= \eta \diag\left(\frac{\vm_{t+1}}{(\vh_{t+1}+e)^2}\right)\left(\beta_2' \mI - (1 - \beta_2') \diag\left(\frac{\nabla f(\vx_t)^{\odot 2}}{\vh_t^2}\right)\right)
\end{align*}

Similarly, at fixed point $\vz_* = [\vx_*, \vm_*, \vh_*]$, we have $\nabla f(\vx_*)^{\odot 2} = \vh_*^{\odot 2}$, this yields:
\[
    \mJ(\vz_*) =
    \begin{bmatrix}
    \frac{\partial \vx_{t+1}}{\partial \vx_t} & -\eta\beta_1 \diag(\frac{1}{\vh_*+e}) & \eta\beta_2 \diag\left(\frac{\vm_*}{(\vh_*+e)^2}\right) \\
    (1 - \beta_1) \nabla^2 f(\vx_*) & \beta_1 \mI & 0 \\
    (1 - \beta_2) \diag(\frac{\nabla f(\vx_*)}{\vh_*}) \nabla^2 f(\vx_*) & 0 & \beta_2 \mI
    \end{bmatrix}.
\]
where
\begin{align*}
    \frac{\partial \vx_{t+1}}{\partial \vx_t} &= (1 - \lambda\eta) \mI - \eta (1 - \beta_1) \diag\left(\frac{1}{\vh_*+e}\right)\nabla^2 f(\vx_*) \\ 
    &\quad + \eta (1 - \beta_2) \diag\left( \frac{\vh_*}{(\vh_*+e)^2} \right) \nabla^2 f(\vx_*)
\end{align*}

Consider the one-dimensional case. The characteristic polynomial of the Jacobian at the fixed point is
\[
\chi_{\mJ}(s) = \det\!\big(s \mI - \mJ(\vz_*)\big)
= s^3 + \alpha_2 s^2 + \alpha_1 s + \alpha_0 ,
\]
where
\begin{align*}
\alpha_2 &= \frac{\eta H_*}{\hbar_*^{2}}\Big[(1-\beta_1)\hbar_* + (\beta_2-1)h_*\Big]
-(\beta_1+\beta_2+1)
+\eta\lambda, \\
\alpha_1 &= \frac{\eta H_*}{\hbar_*^{2}}\Big[\beta_1\beta_2(\hbar_* - h_*) + \beta_1 h_* - \beta_2 \hbar_*\Big]
+(\beta_1\beta_2+\beta_1+\beta_2)
-\eta\lambda(\beta_1+\beta_2), \\
\alpha_0 &= \beta_1\beta_2(\eta\lambda-1).
\end{align*}
To analyze the stability criterion given the polynomial, we can take advantage of the following lemma.
\begin{lemma}[Schur--Jury criterion for cubics]
\label{lem:schur_cubic}
All roots of the cubic polynomial $s^3 + \alpha_2 s^2 + \alpha_1 s + \alpha_0 = 0$ lie strictly inside the unit disk (i.e., the system is Schur stable) if and only if the following four inequalities hold:
\[
|\alpha_0| < 1, \quad
1 + \alpha_2 + \alpha_1 + \alpha_0 > 0, \quad
1 - \alpha_2 + \alpha_1 - \alpha_0 > 0, \quad
1 - \alpha_1 + \alpha_2 \alpha_0 - \alpha_0^2 > 0 .
\]
\end{lemma}

\paragraph{Stability conditions as curvature bounds.}
We express the momentum parameters through the ODE rates $a=(1-\beta_1)/\eta$ and $b=(1-\beta_2)/\eta$, that is, $\beta_1=1-a\eta$ and $\beta_2=1-b\eta$. For each step size $\eta>0$, the map $(\beta_1,\beta_2)\mapsto(a,b)$ is a bijection from $(0,1)^2$ onto $(0,1/\eta)^2$, so Proposition~\ref{prop:discrete_curvature} below covers every choice of $\eta$, $\beta_1$, and $\beta_2$. To compare different step sizes, we hold $a,b$ fixed and let $\beta_1,\beta_2$ vary with $\eta$. This is the setting of the limit $\eta\to0$, which recovers the AdamW ODE, of the dependence of $r_{\mathrm{crit}}$ on $\eta$, and of Corollary~\ref{cor:discrete_counterexamples}. At a fixed point, let
\[
r=|\lambda x_*|,\qquad H_*=f''(x_*),\qquad
\hbar_*=\frac{e}{1-r},\qquad
F''(x_*)=H_*+\frac{\lambda e}{(1-r)^2}.
\]
Define
\begin{align}
A_\eta&=a+\lambda(1-a\eta),\qquad B_\eta=b+\lambda(1-b\eta),\notag\\
C_\eta&=(a+b-ab\eta)A_\eta B_\eta,\notag\\
D_\eta&=a(1-b\eta)A_\eta-rb(1-a\eta)B_\eta,\label{eq:discrete_coefficients}\\
E_\eta&=a(2-b\eta)-rb(2-a\eta),\qquad
T_\eta=(2-\lambda\eta)(2-a\eta)(2-b\eta).\notag
\end{align}
\begin{samepage}
Each coefficient is a polynomial in $\eta$. As $\eta\to0$, $C_\eta$ and $D_\eta$ tend to the constants of the continuous-time condition~\eqref{eq:hurwitz_condition}, whereas $E_\eta$ and $T_\eta$ belong to a condition that only finite steps impose. With these coefficients, the four Jury inequalities reduce to three bounds on the curvature $H_*=f''(x_*)$, an exact discrete counterpart of Proposition~\ref{prop:stable_iff}.

\begin{proposition}[Discrete curvature conditions]\label{prop:discrete_curvature}
Assume $a,b,\lambda,e>0$, let $f$ be $C^2$ near an active one-dimensional fixed point, and suppose
\[
0<\eta<\min\{a^{-1},b^{-1},2\lambda^{-1}\}.
\]
The fixed point is Schur stable if and only if
\begin{align}
F''(x_*)&>0,\label{eq:discrete_minimum}\\
D_\eta H_*&>-\frac{eC_\eta}{1-r},\label{eq:discrete_mismatch}\\
\eta^2E_\eta H_*&<\frac{eT_\eta}{1-r}.\label{eq:discrete_flip}
\end{align}
Thus these conditions ensure local asymptotic stability of the exact nonlinear AdamW update.
\end{proposition}
\end{samepage}

\begin{proof}
Set $p=\beta_1$, $q=\beta_2$, $d=\eta\lambda$, and $k=\eta H_*/\hbar_*$. The characteristic polynomial above factors as
\[
\begin{aligned}
P(s)&=(s-1+d)(s-p)(s-q)\\
&\quad+ks\big[(1-p)(s-q)-r(1-q)(s-p)\big].
\end{aligned}
\]
Its coefficients satisfy
\begin{align*}
|\alpha_0|&=pq|1-d|<1,\\
P(1)&=(1-p)(1-q)\big[d+k(1-r)\big],\\
-P(-1)&=T_\eta-\eta^2E_\eta H_*/\hbar_*,\\
1-\alpha_1+\alpha_2\alpha_0-\alpha_0^2
&=\eta^3\big(C_\eta+D_\eta H_*/\hbar_*\big).
\end{align*}
The first inequality follows from $p,q\in(0,1)$ and $0<d<2$. Since
$d+k(1-r)=\eta e F''(x_*)/\hbar_*^2$, positivity of the remaining three expressions gives the three stated inequalities. Lemma~\ref{lem:schur_cubic} completes the proof.
\end{proof}

\paragraph{What each condition rules out.}
Each condition in Proposition~\ref{prop:discrete_curvature} is the positivity of one Jury expression, and it fails when an eigenvalue of $\mJ(\vz_*)$ leaves the unit disk at a specific point of the unit circle. That point determines how a small perturbation of the fixed point grows (Table~\ref{tab:discrete_conditions}). We name the three bounds accordingly.
\begin{itemize}[leftmargin=*]
\item \textbf{Implicit-minimum floor (\ref{eq:discrete_minimum}).} The fixed point must be a strict local minimum of the implicit objective $F$, exactly as in continuous time (Corollary~\ref{cor:stable_localmin}). Otherwise a real eigenvalue exceeds $+1$, and perturbations grow without changing sign.
\item \textbf{Moment-mismatch bound (\ref{eq:discrete_mismatch}).} This is the discrete form of the Hessian constraint~\eqref{eq:hurwitz_condition}. Its direction depends on the sign of $D_\eta$: it is a floor on $H_*$ when $D_\eta>0$ and a ceiling when $D_\eta<0$. At equality, with the other Jury inequalities strict, a complex-conjugate pair reaches the unit circle, so perturbations rotate while they grow. This is the oscillatory instability caused by unequal moment timescales, consistent with the persistent oscillations in~\figref{fig:fiveplots}.
\item \textbf{Step-flip ceiling (\ref{eq:discrete_flip}).} This bound has no continuous-time counterpart. When $E_\eta>0$, it caps the curvature at $eT_\eta/(\eta^2(1-r)E_\eta)$. Above this cap, an eigenvalue passes $-1$: each step overshoots $x_*$, and the perturbation flips sign from one iteration to the next. This is the analogue of the familiar $\eta<2/L$ step-size limit of gradient descent.
\end{itemize}

\begin{table}[h]
\centering
\caption{The three conditions of Proposition~\ref{prop:discrete_curvature}. Each fails when an eigenvalue of $\mJ(\vz_*)$ crosses the unit circle at the listed point.}
\label{tab:discrete_conditions}
\small
\begin{tabular}{@{}llll@{}}
\toprule
Condition & Crossing & Perturbation if violated & ODE limit ($\eta\to0$) \\
\midrule
Implicit-minimum floor (\ref{eq:discrete_minimum}) & real, at $+1$ & grows monotonically & $F''(x_*)>0$ \\
Moment-mismatch bound (\ref{eq:discrete_mismatch}) & complex pair & grows while oscillating & condition (\ref{eq:hurwitz_condition}) \\
Step-flip ceiling (\ref{eq:discrete_flip}) & real, at $-1$ & flips sign every step & vacuous \\
\bottomrule
\end{tabular}
\end{table}

\paragraph{Recovery of the continuous-time bound.}
As $\eta\to0$ with $a,b$ fixed,
\[
D_\eta\to a(a+\lambda)-rb(b+\lambda),\qquad
C_\eta\to(a+b)(a+\lambda)(b+\lambda)=C,
\]
where $C$ is the constant in~\eqref{eq:hurwitz_condition}. Hence the moment-mismatch bound~\eqref{eq:discrete_mismatch} recovers the continuous-time constraint of Proposition~\ref{prop:stable_iff} exactly. The implicit-minimum floor~\eqref{eq:discrete_minimum} does not depend on $\eta$ at all, since neither the fixed-point equation nor $F$ involves $\eta$, $\beta_1$, or $\beta_2$. The step-flip ceiling~\eqref{eq:discrete_flip} disappears in the limit: its left side is $O(\eta^2)$ while $T_\eta\to8$, so for any fixed $H_*$ it holds once $\eta$ is small enough. Finite step sizes therefore leave the implicit-minimum floor unchanged, make the moment-mismatch bound interact with $\eta$ through $C_\eta$ and $D_\eta$, and add one new obstruction, the step-flip ceiling.

The first two conditions also keep their continuous-time instability modes. Let $\mJ_{\mathrm{ODE}}(\vz_*)$ denote the ODE Jacobian in~(\ref{eq:jacobian}). Substituting $\beta_1=1-a\eta$ and $\beta_2=1-b\eta$ into the discrete Jacobian at the fixed point gives, entrywise,
\[
\mJ(\vz_*)=\mI+\eta\,\mJ_{\mathrm{ODE}}(\vz_*)+O(\eta^2).
\]
For instance, the diagonal entries are $1-\lambda\eta+O(\eta^2)$, $\beta_1=1-a\eta$, and $\beta_2=1-b\eta$, matching $-\lambda$, $-a$, and $-b$ in~(\ref{eq:jacobian}). Each eigenvalue $\mu$ of $\mJ_{\mathrm{ODE}}(\vz_*)$ thus yields an eigenvalue $s=1+\eta\mu+O(\eta^2)$ of $\mJ(\vz_*)$, and $|s|<1$ corresponds to $\Re\mu<0$ for small $\eta$. The implicit-minimum floor fails through a real eigenvalue at $s=+1$, the counterpart of the ODE eigenvalue $\mu=0$; both occur exactly when $F''(x_*)=0$. The moment-mismatch bound fails through a complex pair on the unit circle, the counterpart of the ODE pair crossing the imaginary axis; for small $\eta$ the two crossings agree up to $O(\eta)$, while at larger $\eta$ the crossing angle and the bound itself depend on $\eta$. The step-flip crossing at $s=-1$ has no counterpart, because it requires $\eta\mu\approx-2$, far from $s\approx1$.

\begin{figure}[tb]
    \centering
    \includegraphics[width=\linewidth]{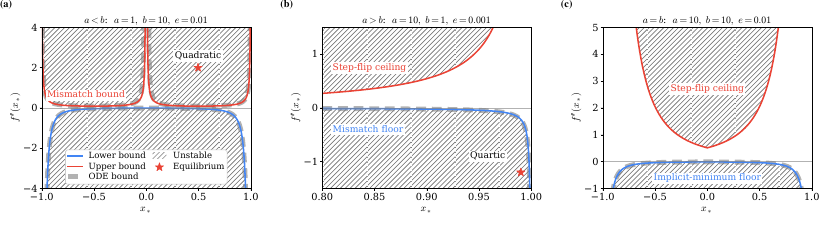}
    \vspace{-1.2em}
    \caption{Schur-stability windows (Proposition~\ref{prop:discrete_curvature}) with $\lambda=1$, $\eta=0.07$, $\beta_1=1-a\eta$, $\beta_2=1-b\eta$. Fixed points are stable between the lower (blue) and upper (red) bounds; hatched regions are unstable, and dashed grey curves are the ODE bounds. Red stars mark the examples of~\secref{sec:nonconverge}. (a)~$a<b$: a moment-mismatch ceiling excludes the convex quadratic. (b)~$a>b$: the moment-mismatch floor (blue, below zero) excludes the quartic, whose $f''(x_*)<0$; for $f''>0$ only the step-flip ceiling applies. (c)~$a=b$: only the implicit-minimum floor and step-flip ceiling remain.}
    \label{fig:discrete_curvature}
    \vspace{-1em}
\end{figure}

\paragraph{Curvature windows in the three momentum regimes.}
Figure~\ref{fig:discrete_curvature} plots, for each fixed-point location $x_*$, the interval of curvatures $f''(x_*)$ for which the fixed point is Schur stable. Unstable curvatures are hatched, and the dashed grey curves show the continuous-time bounds. Since $x_*=0$ is not an active fixed point, the curves there are limits from $x_*\neq0$. We take $\eta=0.07$, larger than the step size $\eta=0.01$ of the examples in~\secref{sec:nonconverge}, so that the effects of the discrete steps are visible. We keep the examples' rates $a,b$, so their momentum parameters differ from those in~\secref{sec:nonconverge} (e.g., $\beta_1=0.3$ and $\beta_2=0.93$ for the quartic). The three panels correspond to the three possible orderings of the momentum rates.
\begin{enumerate}[leftmargin=*]
\item \textbf{Reversed momenta ($a<b$, i.e., $\beta_1>\beta_2$); panel (a).}
The coefficient $D_\eta$ is positive for small $r$ and negative for large $r$. The moment-mismatch bound therefore becomes a ceiling once the fixed point is far enough from the origin:
\begin{equation}\label{eq:discrete_location_threshold}
r>r_{\mathrm{crit}}(\eta)
\coloneqq\frac{a(1-b\eta)A_\eta}{b(1-a\eta)B_\eta}.
\end{equation}
For $r>r_{\mathrm{crit}}(\eta)$, this ceiling lies just above zero (about $0.087$ at the quadratic example, against $0.091$ for the ODE), so even mildly convex minima are unstable. This includes the quadratic, which has $f''=2$. Larger steps make this obstruction slightly more widespread. The threshold $r_{\mathrm{crit}}(\eta)$ decreases from its ODE value $a(a+\lambda)/(b(b+\lambda))$ (here $0.018$, the dashed lines in~\figref{fig:bound}) to $0.006$ at $\eta=0.07$, so the ceiling applies to more fixed-point locations. In general,
\[
\frac{\mathrm d}{\mathrm d\eta}\log r_{\mathrm{crit}}(\eta)
=(b-a)\left[-\frac{1}{(1-a\eta)(1-b\eta)}
             +\frac{\lambda^2}{A_\eta B_\eta}\right]<0,
\]
because $A_\eta B_\eta>\lambda^2(1-a\eta)(1-b\eta)$.

\item \textbf{Usual momenta ($a>b$, i.e., $\beta_1<\beta_2$); panel (b).}
Both $D_\eta$ and $E_\eta$ are positive for $0\le r<1$, so stability requires $H_*$ to lie in the window
\begin{equation}\label{eq:discrete_window}
\max\!\left\{-\frac{\lambda e}{(1-r)^2},
             -\frac{eC_\eta}{(1-r)D_\eta}\right\}
<H_*<\frac{eT_\eta}{\eta^2(1-r)E_\eta}.
\end{equation}
The lower end is the larger of the implicit-minimum floor and the moment-mismatch floor. It nearly coincides with the ODE floor ($-0.215$ versus $-0.224$ at the quartic example), so negative curvature is still excluded even when $F''(x_*)>0$. This floor excludes the quartic, which has $f''(x_*)=-1.2$. The upper end is the step-flip ceiling, which the ODE does not have. It scales as $e/(\eta^2(1-r))$, so it only constrains large steps or sharp minima. To verify the signs, write $S=a+b-ab\eta$, $p=1-a\eta$, and $q=1-b\eta$. At $r=1$, $D_\eta=(a-b)(S+\lambda pq)>0$ and $E_\eta=2(a-b)>0$, and both increase as $r$ decreases.

\item \textbf{Equal momenta ($a=b$); panel (c).}
The moment-mismatch bound is now implied by the implicit-minimum floor, so only the floor and the step-flip ceiling remain:
\begin{equation}\label{eq:discrete_equal_momenta}
-\frac{\lambda e}{(1-r)^2}<H_*<
\frac{e(2-\lambda\eta)(2-a\eta)}{a\eta^2(1-r)^2}.
\end{equation}
Indeed, for $p=q$ the polynomial factors as
\[
P(s)=(s-p)\big[(s-1+d)(s-p)+k(1-p)(1-r)s\big].
\]
The quadratic factor is Schur stable exactly when $d+k(1-r)>0$ and $(2-d)(1+p)>k(1-p)(1-r)$, because its determinant bound holds automatically in the stated step range. These two inequalities give~\eqref{eq:discrete_equal_momenta}. As $\eta\to0$ the ceiling disappears, and stability reduces to $F''(x_*)>0$, recovering Corollary~\ref{cor:vanish} for $a=b$.
\end{enumerate}
In summary, the moment-mismatch obstructions of the continuous-time analysis survive discretization almost unchanged, and finite steps add only the step-flip ceiling on overly sharp minima.

\paragraph{Persistence of the non-convergence examples.}
In Figure~\ref{fig:discrete_curvature}, both main-text examples (red stars) lie outside their stability windows. The next result shows that this does not depend on the choice $\eta=0.07$: with their ODE rates held fixed, both counterexamples remain unstable throughout the admissible step range.
\begin{corollary}\label{cor:discrete_counterexamples}
Consider the quadratic and quartic in~\secref{sec:nonconverge}, with respective settings $(a,b,e,\lambda)=(1,10,0.01,1)$ and $(10,1,0.001,1)$. Hold these ODE rates $a,b$ fixed and let $\beta_1=1-a\eta$ and $\beta_2=1-b\eta$ vary with $\eta$. Then both fixed points are unstable for every $0<\eta<0.1$.
\end{corollary}
\begin{proof}
Let $K=H_*/\hbar_*$. In both cases, $C_\eta\le242$ throughout $0<\eta<0.1$. For the quadratic,
\[
x_*=\frac{3.01-\sqrt{3.01^2-8}}4\in(0.495,0.496),
\qquad K=200(1-x_*)>100.8,
\]
and $D_\eta\le2-0.495\cdot10\cdot0.9\cdot10=-42.55$. Hence
$C_\eta+D_\eta K<242-42.55\cdot100.8<0$.
For the quartic as specified in the main text, $x_*=0.99$, $H_*=-1.2$, and $\hbar_*=0.1$, so $K=-12$. Moreover,
$D_\eta\ge10\cdot0.9\cdot10-0.99\cdot2=88.02$, giving
$C_\eta+D_\eta K\le242-12\cdot88.02<0$.
Thus the oscillation Jury expression is strictly negative in both cases. Since polynomials with all roots in the closed unit disk satisfy the non-strict Jury inequalities, each Jacobian has an eigenvalue of modulus greater than one. The fixed points are therefore unstable.
\end{proof}

\FloatBarrier

\subsection{Multidimensional Case}
\begin{lemma}[Restatement of Lemma~\ref{lem:linear_ode}]
Let $\vz_* = (\vx_*, \vm_*, \vh_*)$ be an equilibrium point of the AdamW ODE~\eqref{eq:adamw_ode}.  
Then the linearized dynamics $\dot{\vz}_t = \mJ(\vz_*) \vz_t$ imply that $\vx_t$ satisfies
\begin{equation}
    \dddot{\vx}_t 
    + \mA_2 \ddot{\vx}_t
    + \mA_1 \dot{\vx}_t
    + \mA_0 \vx_t = 0,
\end{equation}
where
\begin{align*}
    \mA_2 &= (a+b+\lambda) \mI \\
    \mA_1 &= (ab + a\lambda + b\lambda)\mI
            + (a-b)\,\diag(1/(\vh_* + e)) \nabla^2 f(\vx_*)
            + b\,\diag(e/(\vh_* + e)^2) \nabla^2 f(\vx_*), \\
    \mA_0 &= ab\left[\,\diag(e/(\vh_* + e)^2) \nabla^2 f(\vx_*) + \lambda \mI\,\right],
\end{align*}
\end{lemma}

\begin{proof}
Linearize the AdamW ODE~\eqref{eq:adamw_ode} at the equilibrium $\vz_*=(\vx_*,\vm_*,\vh_*)$.
Let $(\vx,\vm,\vh)$ denote infinitesimal perturbations (we drop $\delta$ notation for readability).
Writing $H \coloneqq \nabla^2 f(\vx_*)$ and
\[
U \coloneqq \diag\Big(\frac{1}{\vh_*+e}\Big), \,
V \coloneqq \diag\Big(\frac{\vm_*}{(\vh_*+e)^2}\Big), \,
D \coloneqq \diag\Big(\frac{\nabla f(\vx_*)}{\vh_*}\Big), \,
S \coloneqq \diag\Big(\frac{e}{(\vh_*+e)^2}\Big),
\]
the linearized dynamics are
\begin{equation}\label{eq:lin}
\dot{\vx} = -\lambda \vx - U\vm + V\vh, \quad
\dot{\vm} = a(H\vx - \vm), \quad
\dot{\vh} = b(DH\vx - \vh).
\end{equation}

Differentiate the first equation and substitute \eqref{eq:lin}:
\[
\ddot{\vx}
= -\lambda \dot{\vx} - U\dot{\vm} + V\dot{\vh}
= -\lambda \dot{\vx} - aUH\vx + aU\vm + bVD H\vx - bV\vh.
\]
Using the identity
\[
VD
= \diag\!\Big(\frac{\vm_*}{(\vh_*+e)^2}\Big)\diag\!\Big(\frac{\nabla f(\vx_*)}{\vh_*}\Big)
= \diag\!\Big(\frac{\vh_*}{(\vh_*+e)^2}\Big)
= U - S,
\]
where we used $\vm_*=\nabla f(\vx_*)$ and $\vh_* = \lvert \nabla f(\vx_*) \rvert$ (elementwise) at equilibrium, we obtain
\[
\ddot{\vx} + \lambda \dot{\vx} + (a-b)UH\vx + bSH\vx
= aU\vm - bV\vh.
\]
Eliminate $U\vm$ using the first line of \eqref{eq:lin}:
$U\vm = -\dot{\vx} - \lambda \vx + V\vh$, and thus
\begin{equation}\label{eq:aux2}
\ddot{\vx} + (\lambda + a)\dot{\vx} + (a-b)UH\vx + bSH\vx + a\lambda \vx
= (a-b)V\vh.
\end{equation}
Multiplying \eqref{eq:aux2} by $b$ gives
\begin{equation}\label{eq:Vheq}
b(a-b)V\vh
= b\Big[\ddot{\vx} + (\lambda + a)\dot{\vx} + (a-b)UH\vx + bSH\vx + a\lambda \vx\Big].
\end{equation}

Differentiate \eqref{eq:aux2} and substitute $\dot{\vh}=b(DH\vx-\vh)$ from \eqref{eq:lin}:
\begin{align*}
\dddot{\vx}
&+ (\lambda + a)\ddot{\vx}
+ \big[(a-b)UH + bSH + a\lambda \mI\big]\dot{\vx}
- b(a-b)VD\,H\vx
+ b(a-b)V\vh
= 0.
\end{align*}

Using $VD = U - S$, this becomes
\begin{align*}
\dddot{\vx}
&+ (\lambda + a)\ddot{\vx}
+ \big[(a-b)UH + bSH + a\lambda \mI\big]\dot{\vx}
- b(a-b)(U-S)H\vx
+ b(a-b)V\vh
= 0.
\end{align*}

Substituting \eqref{eq:Vheq} to eliminate $b(a-b)V\vh$ gives
\begin{align*}
\dddot{\vx}
&+ (a+b+\lambda)\ddot{\vx}
+ \Big[(ab + a\lambda + b\lambda)\mI + (a-b)UH + bSH\Big]\dot{\vx}
+ ab\Big[SH + \lambda \mI\Big]\vx
= 0.
\end{align*}

Recognizing $U=\diag(1/(\vh_*+e))$ and $S=\diag(e/(\vh_*+e)^2)$ yields the claimed coefficients
\begin{align*}
    \mA_2 &= (a+b+\lambda) \mI, \\
    \mA_1 &= (ab + a\lambda + b\lambda)\mI
            + (a-b)\,\diag(1/(\vh_* + e)) \nabla^2 f(\vx_*)
            + b\,\diag(e/(\vh_* + e)^2) \nabla^2 f(\vx_*), \\
    \mA_0 &= ab\left[\,\diag(e/(\vh_* + e)^2) \nabla^2 f(\vx_*) + \lambda \mI\,\right],
\end{align*}
which completes the proof.
\end{proof}

\begin{lemma}[First-order reduction]
\label{lem:first_order_reduction}
Let $\vx_t$ solve the third-order ODE
\[
\dddot{\vx}_t + \mA_2 \ddot{\vx}_t + \mA_1 \dot{\vx}_t + \mA_0 \vx_t = 0.
\]
Define $\vy_t \coloneqq \dot{\vx}_t$ and $\vz_t \coloneqq \dot{\vy}_t$. Then $(\vx_t,\vy_t,\vz_t)$ satisfies
\[
\dot{\vx} = \vy, \quad
\dot{\vy} = \vz, \quad
\dot{\vz} = - \mA_0 \vx - \mA_1 \vy - \mA_2 \vz .
\]
\end{lemma}

\begin{proof}
Immediate by substitution: $\dot{\vx}=\vy$, $\dot{\vy}=\vz$, and using $\dddot{\vx}=\dot{\vz}$ in the original ODE.
\end{proof}

\begin{lemma}[Derivative identities]
\label{lem:derivative_identities}
Let $(\vx,\vy,\vz)$ evolve according to Lemma~\ref{lem:first_order_reduction}.
For any symmetric matrices $\mP,\mQ,\mK$ define
\[
W_1(\vx,\vy) \coloneqq \|\mA_2 \vx + \vy\|_{\mP}^2, \quad
W_2(\vy,\vz) \coloneqq \|\mA_2 \vy + \vz\|_{\mQ}^2, \quad
W_3(\vy) \coloneqq \|\vy\|_{\mK}^2,
\]
with $\|u\|_{\mM}^2 \coloneqq u^\top \mM u$. Then
\begin{align*}
\ddt W_1 &= 2(\mA_2 \vx + \vy)^\top \mP (\mA_2 \vy + \vz), \\
\ddt W_2 &= - 2(\mA_2 \vy + \vz)^\top (\mQ \mA_0 \vx + \mQ \mA_1 \vy), \\
\ddt W_3 &= 2\,\vy^\top \mK \vz.
\end{align*}
\end{lemma}

\begin{proof}
For $W_1$, use $\dot{\vx}=\vy$ and $\dot{\vy}=\vz$:
\[
\ddt(\mA_2 \vx + \vy) = \mA_2 \vy + \vz,
\]
hence $\ddt W_1 = 2(\mA_2 \vx + \vy)^\top \mP (\mA_2 \vy + \vz)$.

For $W_2$, note
\[
\ddt(\mA_2 \vy + \vz) = \mA_2 \vz + \dot{\vz} = - \mA_0 \vx - \mA_1 \vy,
\]
so $\ddt W_2 = 2(\mA_2 \vy + \vz)^\top \mQ (- \mA_0 \vx - \mA_1 \vy)$.

For $W_3$, $\dot{\vy}=\vz$, hence $\ddt W_3 = 2\,\vy^\top \mK \vz$.
\end{proof}

\begin{lemma}[Lyapunov cancellation]
\label{lem:algebraic_cancellation}
Let $\mQ$ be symmetric positive definite and define
\[
\mP \coloneqq \mQ \mA_0 \mA_2^{-1}, \quad
\mK \coloneqq \mQ(\mA_1 - \mA_0 \mA_2^{-1}), \quad
\mL \coloneqq \mA_2^\top \mQ(\mA_1 - \mA_0 \mA_2^{-1}).
\]
Assume $\mQ \mA_0$ and $\mQ \mA_1$ are symmetric. Consider
\[
V(\vx,\vy,\vz) \coloneqq \tfrac{1}{2}\Big(\|\mA_2 \vx + \vy\|_{\mP}^2
+ \|\mA_2 \vy + \vz\|_{\mQ}^2 + \|\vy\|_{\mK}^2\Big).
\]
Then along trajectories of Lemma~\ref{lem:first_order_reduction},
\[
\ddt V(\vx_t,\vy_t,\vz_t) = - \tfrac{1}{2}\,\vy^\top(\mL + \mL^\top)\vy.
\]
\end{lemma}

\begin{proof}
Combine Lemma~\ref{lem:derivative_identities}:
\begin{align*}
\ddt V
&= (\mA_2 \vx + \vy)^\top \mP (\mA_2 \vy + \vz)
- (\mA_2 \vy + \vz)^\top (\mQ \mA_0 \vx + \mQ \mA_1 \vy)
+ \vy^\top \mK \vz.
\end{align*}
With $\mP=\mQ \mA_0 \mA_2^{-1}$ we have $\mQ \mA_0 - \mP \mA_2 = 0$ and
$\mQ \mA_1 - \mP = \mK$. Substituting yields
\begin{align*}
\ddt V
&= - (\mA_2 \vy + \vz)^\top \mK \vy + \vy^\top \mK \vz \\
&= - \vy^\top \mA_2^\top \mK \vy \\
&= - \vy^\top \mA_2^\top \mQ(\mA_1 - \mA_0 \mA_2^{-1}) \vy \\
&= - \tfrac{1}{2}\,\vy^\top(\mL + \mL^\top)\vy.
\end{align*}
\end{proof}


\begin{lemma}[Lyapunov stability for the third-order system]
\label{prop:third_order_stability}
Consider $\dddot{\vx}_t + \mA_2 \ddot{\vx}_t + \mA_1 \dot{\vx}_t + \mA_0 \vx_t = 0$.
If there exists a symmetric positive definite $\mQ$ such that
\begin{enumerate}
    \item $\mQ \mA_0$ and $\mQ \mA_1$ are symmetric;
    \item $\mQ \mA_0$, $\mQ \mA_1$, and $\mQ(\mA_1 \mA_2 - \mA_0)$ are positive definite,
\end{enumerate}
then the dynamics are asymptotically stable.
\end{lemma}

\begin{proof}
By Lemma~\ref{lem:first_order_reduction}, stability of the third-order system
is equivalent to stability of its first-order form.
Define $\mP,\mK,\mL$ as in Lemma~\ref{lem:algebraic_cancellation}.
Assumptions (1)–(2) ensure $\mP,\mK \succ 0$ and $\mL+\mL^\top \succ 0$.
Thus $V$ in Lemma~\ref{lem:algebraic_cancellation} is positive definite and radially unbounded.

Along trajectories,
\[
\ddt V(\vx_t,\vy_t,\vz_t) = - \tfrac{1}{2}\,\vy_t^\top(\mL + \mL^\top)\vy_t \le 0,
\]
so $V$ is nonincreasing. Hence trajectories are bounded and Lyapunov stable.

By LaSalle’s invariance principle, the $\Omega$-limit set lies inside
$\mathcal{E}=\{(\vx,\vy,\vz): \vy=0\}$. On $\mathcal{E}$, $\dot{\vx}=0$, so $\vx$ is constant;
also $\dot{\vy}=\vz=0$. Finally $\dot{\vz}=-\mA_0 \vx=0$ implies $\vx=0$ because $\mQ \mA_0 \succ 0$.
Thus the only invariant set is $(0,0,0)$, so the system is asymptotically stable.

In this derivation we used the fact that $\mA_2$ is symmetric positive definite
and commutes with every matrix. This holds automatically because
\[
\mA_2 = (a+b+\lambda)\mI,
\]
a scalar multiple of the identity.
\end{proof}

\section{GPT-2-Small Experiment Configuration}
For the GPT-2-Small experiment in \figref{fig:adamme_gpt_small}, we followed the GPT-opt training setup of \citet{amsel2025polar}. The model is a GPT-2-Small architecture trained from scratch with 12 layers, 12 attention heads, hidden width 768, vocabulary size 50,257, context length 1024, and flash attention. Training uses the FineWeb10B GPT-2-tokenized shards with batch size 512, one epoch, gradient clipping at 1.0, bfloat16 autocast, \texttt{torch.compile}, and a flat-then-linear-decay learning rate schedule: the learning rate stays at its base value for the first 40\% of optimizer steps and then linearly decays to 10\% of the base value. Each optimizer step accumulates $2^{19}$ tokens, and validation is run every 400 optimizer steps on $2^{19}$ tokens. A single tuning run, including the validation passes during training, costs approximately 10 H100 hours. For both AdamW and AdamMe, we tuned learning rate over $\{2\cdot10^{-4},5\cdot10^{-4},10^{-3},2\cdot10^{-3}\}$, weight decay over $\{0.005,0.02,0.1\}$, and optimizer $\epsilon$ over $\{10^{-8},10^{-7},10^{-6},10^{-5}\}$. The best setting used learning rate $5\cdot10^{-4}$ and weight decay $0.1$ for both methods, with $\epsilon=10^{-8}$ for AdamW and $\epsilon=10^{-5}$ for AdamMe.

\paragraph{Asset credit and licenses.}
The training pipeline reuses the publicly available GPT-opt codebase associated with \citet{amsel2025polar},\footnote{\url{https://github.com/gowerrobert/GPT-opt}} with the polar-branch fork.\footnote{\url{https://github.com/modichirag/GPT-opt/tree/polar}} The training corpus is the 10B-token GPT-2-tokenized subset (\texttt{sample/10BT}) of the FineWeb dataset \citep{penedo2024fineweb},\footnote{\url{https://huggingface.co/datasets/HuggingFaceFW/fineweb}} which is released under the Open Data Commons Attribution License (ODC-By) v1.0; we use it in accordance with that license.

\section{The Use of Large Language Models (LLMs)}
We used large language models (LLMs) as a general-purpose assistive tool in preparing this paper. Specifically, LLMs were employed to polish the writing for clarity and readability, and to provide stylistic suggestions. In addition, we interacted with LLMs during the process of generating plots and conducting numerical checks to validate some of the theoretical results, including symbolic and numerical checks of the proposed Lyapunov function for AdamMe during its development. The research questions, theoretical analysis, and main contributions were formulated, derived, and written by the authors.

\clearpage
\section*{NeurIPS Paper Checklist}

\begin{enumerate}

\item {\bf Claims}
    \item[] Question: Do the main claims made in the abstract and introduction accurately reflect the paper's contributions and scope?
    \item[] Answer: \answerYes{}
    \item[] Justification: The two contributions stated in the abstract and Section~\ref{sec:introduction}---the implicit fixed-point objective (Theorem~\ref{thm:objective}) and the stability/non-convergence analysis (Proposition~\ref{prop:stable_iff} and the examples)---are exactly what the paper establishes.

\item {\bf Limitations}
    \item[] Question: Does the paper discuss the limitations of the work performed by the authors?
    \item[] Answer: \answerYes{}
    \item[] Justification: The paper analyzes an idealized continuous-time deterministic regime (defers discrete-time to Appendix~\ref{sec:discrete}), and explicitly notes that the non-convergence examples are worst-case constructions that need not imply detrimental behavior in large-scale training (Section~\ref{sec:conclusions} and discussion after the examples).

\item {\bf Theory assumptions and proofs}
    \item[] Question: For each theoretical result, does the paper provide the full set of assumptions and a complete (and correct) proof?
    \item[] Answer: \answerYes{}
    \item[] Justification: All theorems, propositions, and lemmas state their assumptions explicitly and are numbered and cross-referenced; full proofs are provided in the appendix.

    \item {\bf Experimental result reproducibility}
    \item[] Question: Does the paper fully disclose all the information needed to reproduce the main experimental results of the paper to the extent that it affects the main claims and/or conclusions of the paper (regardless of whether the code and data are provided or not)?
    \item[] Answer: \answerYes{}
    \item[] Justification: The main results are theoretical and are supported by deterministic low-dimensional simulations; the objectives, optimizer hyperparameters $(\eta,\beta_1,\beta_2,\lambda,e)$, and equilibria are stated in the figure captions and examples. For the GPT-2-Small run, the model, data, schedule, validation protocol, tuning grid, and selected hyperparameters are summarized in Appendix~B.

\item {\bf Open access to data and code}
    \item[] Question: Does the paper provide open access to the data and code, with sufficient instructions to faithfully reproduce the main experimental results, as described in supplemental material?
    \item[] Answer: \answerYes{}
    \item[] Justification: We provide AdamMe reference code; Appendix~B specifies the public GPT-opt/FineWeb10B setup.

\item {\bf Experimental setting/details}
    \item[] Question: Does the paper specify all the training and test details (e.g., data splits, hyperparameters, how they were chosen, type of optimizer) necessary to understand the results?
    \item[] Answer: \answerYes{}
    \item[] Justification: All optimizer hyperparameters $(\eta,\beta_1,\beta_2,\lambda,e)$ used in the deterministic figures are listed in the captions and example statements. Appendix~B gives the GPT-2-Small model size, dataset, schedule, validation protocol, optimizer grid, and best AdamW/AdamMe hyperparameters.

\item {\bf Experiment statistical significance}
    \item[] Question: Does the paper report error bars suitably and correctly defined or other appropriate information about the statistical significance of the experiments?
    \item[] Answer: \answerNo{}
    \item[] Justification: The deterministic ODE and iteration trajectories have no sampling variability. The GPT-2-Small experiment is reported as a single run without error bars or significance tests; this is common for LLM pretraining comparisons because repeated runs are computationally expensive, and we do not use this experiment as statistical evidence for the paper's main theoretical claims.

\item {\bf Experiments compute resources}
    \item[] Question: For each experiment, does the paper provide sufficient information on the computer resources (type of compute workers, memory, time of execution) needed to reproduce the experiments?
    \item[] Answer: \answerYes{}
    \item[] Justification: The deterministic numerical illustrations run in seconds on a standard laptop CPU. The GPT-2-Small experiment follows the GPU training setup described in Appendix~B; each reported run uses the stated token budget, batch size, sequence length, mixed precision, validation cadence, and approximately 10 H100 hours per tuning run.
    
\item {\bf Code of ethics}
    \item[] Question: Does the research conducted in the paper conform, in every respect, with the NeurIPS Code of Ethics \url{https://neurips.cc/public/EthicsGuidelines}?
    \item[] Answer: \answerYes{}
    \item[] Justification: The work is purely theoretical with no human subjects, data collection, or deployed system, and conforms to the NeurIPS Code of Ethics.

\item {\bf Broader impacts}
    \item[] Question: Does the paper discuss both potential positive societal impacts and negative societal impacts of the work performed?
    \item[] Answer: \answerNA{}
    \item[] Justification: This is foundational theoretical work on optimizer dynamics with no direct societal application or deployment path.
    
\item {\bf Safeguards}
    \item[] Question: Does the paper describe safeguards that have been put in place for responsible release of data or models that have a high risk for misuse (e.g., pre-trained language models, image generators, or scraped datasets)?
    \item[] Answer: \answerNA{}
    \item[] Justification: The paper releases no data, pretrained models, or other artifacts with misuse risk.

\item {\bf Licenses for existing assets}
    \item[] Question: Are the creators or original owners of assets (e.g., code, data, models), used in the paper, properly credited and are the license and terms of use explicitly mentioned and properly respected?
    \item[] Answer: \answerYes{}
    \item[] Justification: The GPT-opt codebase and FineWeb-10B dataset used in the GPT-2-Small experiment are credited and their license status is documented in Appendix~B.

\item {\bf New assets}
    \item[] Question: Are new assets introduced in the paper well documented and is the documentation provided alongside the assets?
    \item[] Answer: \answerYes{}
    \item[] Justification: We release AdamMe reference code; Algorithm~\ref{alg:adam_variants} documents the optimizer.

\item {\bf Crowdsourcing and research with human subjects}
    \item[] Question: For crowdsourcing experiments and research with human subjects, does the paper include the full text of instructions given to participants and screenshots, if applicable, as well as details about compensation (if any)?
    \item[] Answer: \answerNA{}
    \item[] Justification: The paper involves no crowdsourcing or human subjects.

\item {\bf Institutional review board (IRB) approvals or equivalent for research with human subjects}
    \item[] Question: Does the paper describe potential risks incurred by study participants, whether such risks were disclosed to the subjects, and whether Institutional Review Board (IRB) approvals (or an equivalent approval/review based on the requirements of your country or institution) were obtained?
    \item[] Answer: \answerNA{}
    \item[] Justification: The paper involves no human subjects research.

\item {\bf Declaration of LLM usage}
    \item[] Question: Does the paper describe the usage of LLMs if it is an important, original, or non-standard component of the core methods in this research? Note that if the LLM is used only for writing, editing, or formatting purposes and does \emph{not} impact the core methodology, scientific rigor, or originality of the research, declaration is not required.
    \item[] Answer: \answerNA{}
    \item[] Justification: LLMs were used only for writing polish and auxiliary visualization checks, not as a component of the core methodology; this is disclosed in the appendix (``The Use of Large Language Models'').

\end{enumerate}

\end{document}